\documentclass{article}
\PassOptionsToPackage{numbers, compress}{natbib}
\usepackage[preprint]{neurips_2026}

\usepackage[utf8]{inputenc}
\usepackage[T1]{fontenc}
\usepackage{hyperref}
\usepackage{url}
\usepackage{booktabs}
\usepackage{amsfonts}
\usepackage{amsmath}
\usepackage{amssymb}
\usepackage{nicefrac}
\usepackage{microtype}
\usepackage{xcolor}
\usepackage{graphicx}
\usepackage{subcaption}
\usepackage{multirow}
\usepackage{wrapfig}

\newcommand{\method}{PAMF}
\newcommand{\wmissing}{within-modality missing}
\newcommand{\mmissing}{modality-level missing}
\newcommand{\EncImp}{h_{\mathrm{imp}}}
\newcommand{\EncFus}{h_{\mathrm{fus}}}
\newcommand{\ffm}{f_{\mathrm{fm}}}
\newcommand{\fds}{f_{\mathrm{ds}}}
\newcommand{\Lds}{\mathcal{L}_{\mathrm{ds}}}
\newcommand{\Lfm}{\mathcal{L}_{\mathrm{fm}}}

\title{PAMF: Prior-Aware Multimodal Fusion for Incomplete Time Series Data
} 

\author{%
  Ziwen Kan \and
  Wugeng Zheng \and
  Tianlong Chen$^{2}$ \and
  Song Wang$^{1}$ \\
  $^{1}$Department of Computer Science, University of Central Florida \\
  $^{2}$Department of Computer Science, University of North Carolina at Chapel Hill
}

\begin{document}

\maketitle

\begin{abstract}
In healthcare, multimodal time series tasks often operate on incomplete observations in practice, for example when ECG segments are lost because electrodes detach or an entire respiratory channel is unavailable during overnight monitoring. 
Such missingness typically appears in two structurally distinct patterns: within-modality missing, where values are absent within an otherwise observed modality, and modality-level missing, where an entire modality is unavailable. 
Existing methods typically represent unobserved data implicitly through masks or missing embeddings, without learning instance-specific missing information, and most are designed for only one missingness pattern. 
A natural approach is to explicitly estimate the missing data; however, existing imputation methods treat missingness uniformly despite their different structural priors, and the imputation process is often isolated from downstream tasks, preventing downstream tasks from guiding imputation toward more informative representations.
To address these limitations, we present PAMF, a multimodal time-series framework that explicitly handles different missingness patterns while coupling imputation with downstream prediction through prior-aware flow matching and weight sharing. 
Specifically, the method initializes the flow-matching source state with type-specific priors to distinguish two missing types. 
It further connects imputation and classification through architecturally matched encoders with weight sharing, transferring task-relevant representations into the imputation process.
Experiments on multiple multimodal healthcare time-series benchmarks show that the proposed method achieves the strongest overall downstream performance across diverse datasets and missing settings compared with existing baselines. 
\end{abstract}

\section{Introduction}
\label{sec:intro}

Multimodal time series are central to healthcare
applications~\cite{zhang2023mists, shukla2021mtand, soenksen2022integrated}
and beyond~\cite{tsai2019mult, hoffmann2020review},
where heterogeneous sensors jointly capture various information like different trends and periodic patterns.
~\cite{liu2025mm4tsa, liang2024foundation}.
%
However, in real-world
deployment, these benefits are often limited by pervasive missing data.
For example, in sleep staging from overnight polysomnography, it is often difficult to obtain complete physiological recordings for every subject-night.
Missing data takes two structurally distinct
forms: \textit{\wmissing{}}, where individual time points
are absent within an otherwise recorded modality, for example when a
sensor goes offline for a short period of time, and
\textit{\mmissing{}}, where an entire modality goes
unrecorded~\cite{little2019statistical, wu2026mlmm,
han2024fusemoe}, for example when a patient is not in a
condition to undergo a particular examination such as electrocardiogram (ECG) or
respiratory signal.
If
not handled properly, both types of missingness can seriously degrade
downstream performance. 
How to effectively handle both types of missingness
therefore remains a critical open problem in multimodal
time series analysis~\cite{wu2026mlmm, liu2025mm4tsa, wang2024tsimputation_survey}. 

Early multimodal time series
studies assume all modalities are fully observed and
readily aligned~\cite{tsai2019mult, zhang2023mists,
zadeh2017tfn, rahman2020mag}, limiting their applicability
under real-world missingness.
%
Recently, few methods that do account for missing data take a
conservative approach: either substituting absent
modalities with learned representations~\cite{,
mohapatra2025maestro} or routing around missing inputs via
gating and masks~\cite{han2024fusemoe}, or memory banks~\cite{yun2024flexmoe}, without
learning instance-specific representations for the missing
content, and most such methods address only one of the two
missing types, limiting their applicability in realistic
deployment scenarios.
%
To the best of our knowledge, existing methods do not
simultaneously tackle both challenges. Table~\ref{tab:comparison}
summarizes how they fall short across the key
capabilities required.

A natural way to close this gap is to explicitly reconstruct missing data to address both missing types and generate instance-specific representation. Yet despite this need, no imputation solution is proposed, which we attribute to two independent technical
challenges.
%
First, existing diffusion-based generative imputation
methods~\cite{tashiro2021csdi, alcaraz2023sssd, yang2024fgti}
start from the same domain-agnostic source distribution
$\mathcal{N}(\mathbf{0}, \mathbf{I})$ for all missing cases,
without distinguishing the different structures of
\wmissing{} and \mmissing{}.
%
As a result, the generative process may not be tailored to the specific missing type it must handle.
%
Second, imputation and downstream prediction have largely
been treated as independent, sequential tasks~\cite{ wang2024tsimputation_survey}:
the imputation
module is trained separately from the downstream task, so
task-relevant representations learned later for fusion and downstream
task cannot be used to guide imputation, leaving the
imputation process largely task-agnostic.
%
Consequently, the recovered representations may not be
adapted to the needs of the downstream task.

To address these challenges, we propose \method{}, a unified
framework for multimodal time series analysis under
missingness.
%
For the first challenge, \method{} features a newly designed
flow-matching-based~\cite{lipman2023flow, liu2023rectified}
generative module with \textit{prior-aware initialization},
so that \wmissing{} and \mmissing{} can be handled with
different initial states that reflect their distinct
structural information.
For \wmissing{}, the initialization is constructed from
temporal neighbor and mean values to keep the starting point
within a plausible local observation range, following the intuition of classical carry-forward imputation~\cite{molenberghs2007missing}.
For \mmissing{}, the initialization is derived from other
observed data to provide a starting point that better
matches the underlying distribution. This design also
reduces inference overhead to fewer function evaluations
than diffusion-based alternatives~\cite{tashiro2021csdi, alcaraz2023sssd} require.
%
For the second challenge, we introduce a weight-sharing
mechanism, through which imputation and the downstream task
are connected by transferring representations learned for
downstream prediction into the imputation process.
%
The resulting imputed signals are then used for downstream
prediction to improve downstream performance.
%
Our main contributions are as follows:

\begin{itemize}
    \item We propose \method{}, a unified framework for
    multimodal time series analysis under missingness, with
    \textit{prior-aware initialization} in a flow-matching-based
    generative module, so that \wmissing{} and \mmissing{} can
    be handled with different initial states that reflect their
    distinct structural information.

    \item We introduce a \textit{weight-sharing mechanism} that connects imputation with the downstream task by initializing the imputation branch from representations learned for downstream prediction, allowing task-relevant structure to guide the imputation process.

    \item We evaluate \method{} on four datasets under
    multiple missingness settings, including \wmissing{},
    \mmissing{}, and additional mixed/high-missingness cases,
    demonstrating the strongest overall downstream performance.
\end{itemize}

\begin{table*}[t]
\centering
\caption{Comparison of existing methods against \method{}.
  Methods include MulT~\cite{tsai2019mult}, TFN~\cite{zadeh2017tfn},
  MAG~\cite{rahman2020mag}, FuseMoE~\cite{han2024fusemoe},
  FlexMoE~\cite{yun2024flexmoe}, Maestro~\cite{mohapatra2025maestro},
  MIRA~\cite{li2025mira}, and representative single-modality imputation
  models such as CSDI~\cite{tashiro2021csdi} and
  SSSD~\cite{alcaraz2023sssd}. \textbf{Imp.}: explicit reconstruction;
  \textbf{E2E}: imputation and fusion trained end-to-end.}
\label{tab:comparison}
\vspace{.5em}
\small
\setlength{\tabcolsep}{14pt}
\begin{tabular}{lcccc}
\toprule[1pt]
\textbf{Method} & Within  Missing & Modality Missing & \textbf{Imp.} & \textbf{E2E} \\
\midrule
MulT, TFN, MAG
  & $\times$ & $\times$ & $N/A $ & $ N/A $ \\
FuseMoE
  & $\times$ & \checkmark & $\times$ & $N/A$ \\
FlexMoe & $\times$ & \checkmark & $\times$ & $N/A$ 
\\
Maestro& $\times$ & \checkmark & $\times$ & $N/A$ 
\\
MIRA
  & \checkmark & $\times$ & $\times$ & $N/A$ \\

\midrule
Single-Modality Imputation
  & \checkmark & $\times$ & \checkmark & $\times$ \\ 
\midrule
\textbf{\method{} (Ours)}
  & \checkmark & \checkmark & \checkmark & \checkmark \\
\bottomrule[1pt]
\end{tabular}
\end{table*}

\section{Related Work}
\label{sec:related}

\paragraph{Multimodal Time Series Fusion.}
Multimodal time-series fusion methods typically assume complete
modalities.
%
Tensor fusion networks~\cite{zadeh2017tfn}, multimodal
transformers~\cite{tsai2019mult}, and multimodal adaptation
gates~\cite{rahman2020mag} demonstrate strong gains on
sentiment and affective benchmarks; in the healthcare domain,
MISTS~\cite{zhang2023mists}
extends this line to irregularly sampled electronic health records via
attention-based modality alignment.
%
Although achieving promising results, the assumption of full observation limits their use in realistic clinical settings where missingness is ubiquitous~\cite{chowdhry2021missing,national2011prevention}.

More recent studies tried to tackle this issue by using masks,
routing, or learned placeholders to implicitly represent missing, in line
with a broader missing-modality multimodal learning literature that
focuses on representation robustness rather than explicit signal
recovery~\cite{wu2026mlmm, wang2023shaspec, wu2024muse, yi2024robustmm, dai2025md2n}.
%
For \mmissing{}, FuseMoE~\cite{han2024fusemoe} routes inputs through
a mixture-of-experts gating mechanism preventing the model from attending to missing-modality positions in the cross-encoder.;
Flex-MoE~\cite{yun2024flexmoe} substitutes absent modalities with a
learnable embedding bank, providing fixed representations that do not
condition on the specific observed context;
Maestro~\cite{mohapatra2025maestro} applies adaptive sparse attention
over partially observed streams, yet similarly leaves missing signals unaware.
For \wmissing{}, MIRA~\cite{li2025mira} handles irregular sampling and
missing values through channel-independent sparse mixture-of-experts
modeling, but remains prediction-oriented rather than an explicit
cross-modal signal imputer. Related multimodal physiological and cross-modal generation studies also
emphasize robustness representation under incomplete
observations~\cite{fu2025physioomni, lee2024biofame},
but they do not target the explicit imputation of missing.
%

\paragraph{Time Series Imputation.}
General time-series imputation primarily targets \wmissing{} within
observed channels, as summarized in recent
surveys~\cite{wang2024tsimputation_survey}.
%
Earlier approaches include neural imputers such as
M-RNN~\cite{yoon2018estimating},
BRITS~\cite{cao2018brits}, SAITS~\cite{du2023saits}, and
MVI~\cite{bansal2021missing}
propagate observations across missing positions through recurrent or
attention-based mechanisms;
generative models
such as GP-VAE~\cite{fortuin2020gpvae}, which places a Gaussian
process prior over latent trajectories, and GAN-based
models~\cite{luo2018multivariate}, which learn the data
distribution adversarially.
Motivated by denoising diffusion models~\cite{ho2020denoising},
recent probabilistic imputers model conditional distributions over
missing values more directly and then achieved better imputation results:
CSDI~\cite{tashiro2021csdi} performs conditional score generation
by treating observed time steps as conditioning signals via
two-channel attention;
SSSD~\cite{alcaraz2023sssd} further incorporates structured state
space models for long-range dependency modeling to estimate missingness; and
LSCD~\cite{fons2025lscd} conditions diffusion on frequency-domain
structure for time-series imputation. Related probabilistic
formulations include spatiotemporal diffusion in
PriSTI~\cite{liu2023pristi}, Schr\"odinger-bridge-based
imputation~\cite{chen2023csbi}, consistent diffusion
imputation~\cite{zhou2024mtsci}, and optimal-transport-based
imputation~\cite{wang2025pswi}.
%
These methods improve signal-level imputation within observed
multivariate series, but do not directly address multimodal
\mmissing{}. 
Unlike diffusion-based generation, which simulates stochastic
denoising dynamics, flow matching learns a velocity field
parameterized by the transport variable $\tau$ that defines an ODE transporting samples from a source
distribution to the data distribution.

\paragraph{Flow Matching for Imputation.}
Flow matching~\cite{lipman2023flow, liu2023rectified} therefore offers
a natural alternative for generative imputation.
Compared with diffusion-based iterative denoising, it reduces inference step to as few as
10 to 30 ODE steps and leaves the initial state $x_0$ designable,
making prior-aware initialization practical.
%
Recent imputation-oriented extensions follow the same direction:
CLWF~\cite{qian2024clwf} studies Lagrangian Wasserstein flows for imputation;
MIRI~\cite{yu2025miri} frames imputation as minimizing data-mask
mutual information under a rectified-flow objective; and
CFMI~\cite{simkus2025cfmi} studies conditional flow matching for
general missing-data imputation.
Although these methods differ in path design and conditioning
objectives, they share the same practical premise: the source
distribution can encode useful structure before transport rather than
starting from fixed noise. This makes flow matching particularly
attractive for imputation, where observed temporal context or distribution information can inform initialization.
%
However, existing flow-matching imputation methods still focus on tabular or
unimodal missing-value recovery; they do not encode type-specific
priors for \wmissing{} and \mmissing{} or couple signal-level
imputation with downstream classification.

\section{Method}
\label{sec:method}

\subsection{Problem Formulation}
\label{sec:problem}

For a sample with $M$ modalities, modality $m$ provides a time series
$\mathbf{x}^{(m)} \in \mathbb{R}^{T \times C_m}$,
where $T$ is the shared sequence length and $C_m$ is the number of channels.
%
For each modality $m$, let $\mathbf{R}^{(m)} \in \{0,1\}^{T \times C_m}$ be the
binary observation mask, with value $1$ for observed entries and $0$ for
missing entries.
Let $\Omega_w$ denote the set of \wmissing{} positions, and let
$\Omega_m$ denote the set of \mmissing{} positions, where an entire
modality is absent and thus $\mathbf{R}^{(m)}=\mathbf{0}_{T\times C_m}$.
We write $\Omega=\Omega_w\cup\Omega_m$ for the full set of positions to
impute. 
%
Given partially observed multimodal data and a downstream label $y$,
\method{} initializes missing positions in $\Omega_w$ and $\Omega_m$
with type-specific priors, transports these source states toward
completed signals via flow matching, and then learns task-relevant
multimodal representations from the imputed data for downstream
prediction.

\subsection{Framework Overview}
\label{sec:overview}

\begin{figure*}[t]
  \centering
  \begin{subfigure}[t]{0.82\linewidth}
    \centering
    \includegraphics[width=\linewidth]{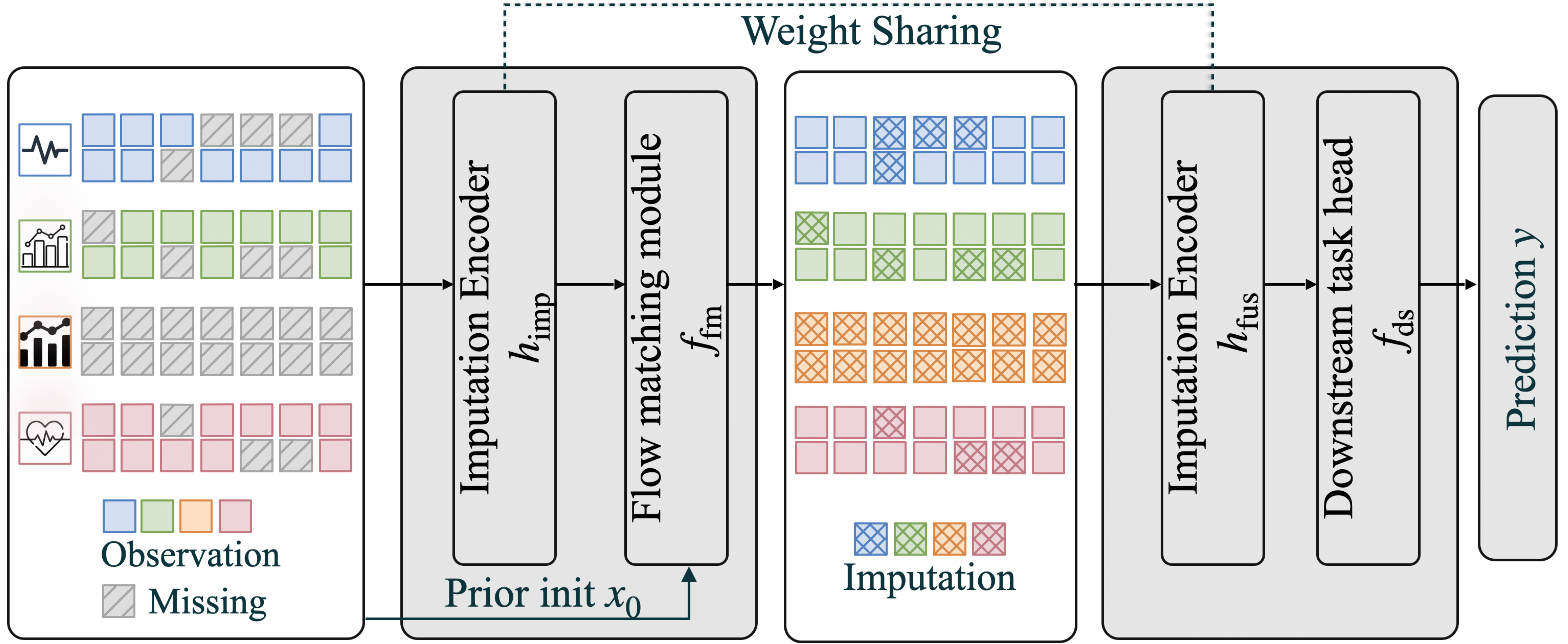}
  \end{subfigure}
  \hfill
  \begin{subfigure}[t]{0.17\linewidth}
    \centering
    \includegraphics[width=\linewidth]{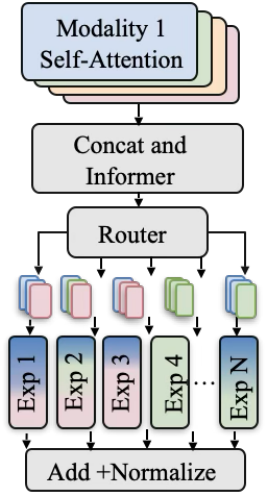}
  \end{subfigure}
  \caption{Overview of \method{}.
    Left: the end-to-end pipeline combines an imputation encoder $\EncImp$,
    a fusion encoder $\EncFus$, a flow matching module $\ffm$, and a downstream
    head $\fds$.
    $\EncImp$ extracts the imputation context
    $\mathbf{c}_{\mathrm{imp}}$ from the observed inputs. Conditioned on
    $\mathbf{c}_{\mathrm{imp}}$ and a prior-aware source state $x_0$,
    $\ffm$ imputes the missing signals to generate completed data, which are then fed into
    $\EncFus$ to produce the fused representation
    $\mathbf{c}_{\mathrm{fus}}$ for downstream head $\fds$ to predict
    $y$. Right: the shared MoE encoder backbone used by both encoders.}
  \label{fig:overview}
\end{figure*}

\method{} addresses imputation and downstream prediction in a single
end-to-end framework built from four components:
an imputation encoder $\EncImp$,
a fusion encoder $\EncFus$,
a flow matching module $\ffm$,
and a task-specific downstream head $\fds$.
%
$\EncImp$ and $\EncFus$ share the same encoder structure.
Concretely, each modality is first linearly projected and encoded by
self-attention; the resulting modality embeddings are then concatenated
and passed to a cross-modal MoE block, following sparsely gated expert
routing~\cite{shazeer2017outrageously} and prior multimodal studies~\cite{han2024fusemoe}, with Informer
attention~\cite{zhou2021informer} to produce the fused context.
Both encoders are designed to learn aligned multimodal
representations and are coupled through three-phase staged training.
In practice, $\EncImp$ provides the context representation
$\mathbf{c}_{\mathrm{imp}}$ used by the imputation branch, while
$\EncFus$ provides the fused representation
$\mathbf{c}_{\mathrm{fus}}$ for downstream prediction.
%
$\ffm$ imputes missing time series values by transporting a
structured source state $x_0$ to the target data distribution
along straight ODE paths~\cite{lipman2023flow, liu2023rectified},
using a conditional velocity network driven by the observed signal and
$\mathbf{c}_{\mathrm{imp}}$.
Instead of fixing $x_0$ to a Gaussian distribution, we initialize
it via \textit{prior-aware initialization} that injects
type-specific structural priors for \wmissing{} and \mmissing{}
respectively (Section~\ref{sec:prior}).
%
$\fds$ is a task-specific downstream head applied to the fused representation
$\mathbf{c}_{\mathrm{fus}}$ to predict the final target $y$.
Due to page limits, detailed implementation details are deferred to
Appendix~\ref{app:impl}.

\subsection{Prior-Aware Flow Matching}
\label{sec:prior}



The intuition for different prior is that starting state for imputation should depend on the missing type.
For \wmissing{}, nearby observed values in the same channel provide a
local temporal prior. For \mmissing{}, same-modality evidence is
absent, so initialization may instead rely on dataset distribution information
from the observation.
Flow matching~\cite{lipman2023flow, liu2023rectified} is well suited to encode
this asymmetry because, unlike diffusion models with a fixed Gaussian start, it
lets us design the source state $x_0$ while transporting it to the target
signal $x_1$ through a continuous vector field.
Here
$x_\tau$ denote the source, target, and intermediate values on those positions.
To let the generative path use information from the observed inputs, we
condition the vector field on the imputation context
$\mathbf{c}_{\mathrm{imp}}$ produced by $\EncImp$. Let
$\Phi_\tau^\theta(\cdot;\mathbf{c}_{\mathrm{imp}})$ denote the resulting flow
map induced by $v_\theta$. The conditional transport process can be written as
\begin{equation}
  \begin{aligned}
    p_\theta(x_1 \mid \mathbf{c}_{\mathrm{imp}})\,
    &=
    \left[\Phi_1^\theta(\cdot;\mathbf{c}_{\mathrm{imp}})\right]_{\#}
    p_0(x_0 \mid \mathbf{c}_{\mathrm{imp}}, \Omega), \\
    \frac{d\Phi_\tau^\theta(x_0;\mathbf{c}_{\mathrm{imp}})}{d\tau}\,
    &=
    v_\theta\!\left(\Phi_\tau^\theta(x_0;\mathbf{c}_{\mathrm{imp}}),\,
    \tau,\, \mathbf{c}_{\mathrm{imp}}\right).
  \end{aligned}
  \label{eq:conditional_transport}
\end{equation}
Here $p_0(x_0 \mid \mathbf{c}_{\mathrm{imp}}, \Omega)$ denotes the source
initialization law, and $[\cdot]_{\#}$ is the push-forward operator.
Equivalently, the state $x_\tau=\Phi_\tau^\theta(x_0;\mathbf{c}_{\mathrm{imp}})$
follows the conditional ODE
\begin{equation}
  \frac{dx_\tau}{d\tau} = v_\theta\!\left(x_\tau,\, \tau,\, \mathbf{c}_{\mathrm{imp}}\right),
  \quad \tau \in [0, 1],
  \label{eq:fm_ode}
\end{equation}
with the boundary condition $x_{\tau=0}=x_0$ and $x_{\tau=1}\approx x_1$.
Here $\mathbf{c}_{\mathrm{imp}}$ plays the standard role of conditioning the
vector field at each FM step.

Following conditional flow matching~\cite{lipman2023flow, tong2024improving},
we train this vector field by regressing to the conditional velocity along a
chosen probability path.
For the optimal-transport linear bridge used in rectified
flows~\cite{lipman2023flow, liu2023rectified}, the path and target velocity
$u_\tau$ are
\begin{equation}
  x_\tau = (1 - \tau)\,x_0 + \tau\,x_1,
  \qquad
  u_\tau(x_\tau \mid x_0,x_1,\mathbf{c}_{\mathrm{imp}})=x_1-x_0,
  \qquad
  \tau \sim \mathcal{U}[0,1].
  \label{eq:fm_path}
\end{equation}
This yields the masked conditional FM regression objective
\begin{equation}
  \Lfm
  =
  \mathbb{E}_{\tau,\,x_0,\,x_1}\!\left[
    \frac{1}{|\Omega|}
    \sum_{i\in\Omega}
    \left\|
    v_\theta\!\left((1-\tau)x_0[i]+\tau x_1[i],\,
    \tau,\,\mathbf{c}_{\mathrm{imp}}\right)
    -
    \left(x_1[i]-x_0[i]\right)
    \right\|_2^2
  \right].
  \label{eq:conditional_fm_loss}
\end{equation}
Thus $\mathbf{c}_{\mathrm{imp}}$ conditions the vector field at every FM step,
while $\Omega=\Omega_w\cup\Omega_m$ restricts supervision to the missing entries.

Our main contribution is to condition the source state itself on the missing
type.
Generalized CFM does not require the source distribution to be Gaussian~\cite{tong2024improving}, which lets us replace an
uninformative Gaussian source with a structured imputation prior.
In practice, we construct $x_0$ directly with missing-type-specific rules.
For each missing position $i=(t,c)$, we define
\begin{equation}
  x_0[i] =
  \begin{cases}
    x_0^{\mathrm{wm}}[i], & i \in \Omega_w, \\[3pt]
    x_0^{\mathrm{ml}}[i], & i \in \Omega_m,
  \end{cases}
  \qquad
  p_0(x_0 \mid \mathbf{c}_{\mathrm{imp}},\Omega)
  =
  p_0^{\mathrm{wm}}(x_0^{\mathrm{wm}}\mid \Omega_w)\,
  p_0^{\mathrm{ml}}(x_0^{\mathrm{ml}}\mid \mathbf{c}_{\mathrm{imp}},\Omega_m).
  \label{eq:prior_aware}
\end{equation}
Thus, \wmissing{} entries use a temporal-neighbor prior, whereas \mmissing{}
entries use a modality-level prior conditioned on
$\mathbf{c}_{\mathrm{imp}}$.
The former exploits local continuity within the same channel, while the latter
relies on information inferred from the present modalities.
Both strategies reduce expected transport distance relative to a
Gaussian start, consistent with the theoretical analysis
in~\cite{liu2023rectified, kollovieh2025tsflow}.
In our implementation, the default \wmissing{} prior is a rule-based temporal
neighbor interpolation,
\begin{equation}
  x_0^{\mathrm{wm}}[t,c]
  = \alpha_{t,c}\, x^{\mathrm{nbr}}_{t,c}
  + (1 - \alpha_{t,c})\, \bar{x}_c^{\mathrm{batch}},
  \qquad
  \alpha_{t,c} = \exp\!\left(-\lambda \cdot |t^* - t|\right),
  \label{eq:prior_wm_main}
\end{equation}
where $t^*$ denotes the time index of the nearest observed entry to $t$ in
channel $c$, so that $x^{\mathrm{nbr}}_{t,c}=x_{t^*,c}$, and
$\bar{x}_c^{\mathrm{batch}}$ is the per-channel batch mean. Here $\lambda > 0$
is a fixed decay coefficient that controls how quickly the neighbor weight
decreases as the temporal gap $|t^*-t|$ grows.
We use this weighting because missing entries are more likely to follow
their nearest temporal neighbor when the gap is short, whereas the global
channel mean becomes a safer fallback as the gap widens.
In the default implementation, we set $\lambda = 0.1$, yielding an exponential
decay of neighbor confidence with temporal distance.
The default \mmissing{} prior is the per-channel batch mean:
\begin{equation}
  x_0^{\mathrm{ml}} = \bar{x}^{\mathrm{batch}}.
  \label{eq:prior_ml_mean_main}
\end{equation}
This choice reflects the intuition that when an entire modality is
absent, a conservative global statistic may provide a stable
initialization that stays close to the typical value range of that
channel before cross-modal conditioning refines it.
We also evaluated richer variants, including a learned-neighbor weighting for
$x_0^{\mathrm{wm}}$ that adapts the neighbor confidence, a batch-matching prior
for $x_0^{\mathrm{ml}}$ that retrieves the most context-similar observed
sample, and a memory-bank prior for $x_0^{\mathrm{ml}}$ that uses learnable
cross-sample prototypes; Appendix~\ref{app:prior} gives their full
formulations, and Table~\ref{tab:prior_ablation} shows that they do not
consistently improve over the simpler defaults above.

\subsection{Weight Sharing and Three-Phase Training}
\label{sec:curriculum}

We design $\EncImp$ and $\EncFus$ with the same encoder architecture. Both imputation and downstream prediction require temporal
within-modality representation and cross-modal correlations, so the two
branches could learn aligned rather than fully separate
representations. Matched encoder structures therefore make
$\mathbf{c}_\mathrm{imp}$ and $\mathbf{c}_\mathrm{fus}$ easier to align
and support knowledge transfer across training stages. Because their learning objective and supervision signals differ, we couple them through a three-phase
training strategy instead of a single joint loss.

Accordingly, the three phases use a downstream task loss
and a flow matching velocity regression loss.
Here $\mathbf{x}_{\mathrm{obs}}$ is the incomplete input, and
$\mathbf{x}_{\mathrm{complete}}$ is the same input after FM imputation:
\begin{align}
  \Lds &=
    \mathrm{CE}\!\left(\fds\!\left(\EncFus(\mathbf{x}_{\mathrm{obs}})\right),\, y\right),
    \label{eq:loss_ds} \\
  \Lfm &=
    \mathbb{E}_{\tau \sim \mathcal{U}[0,1]}\!\left[
      \left\| v_\theta\!\left(x_\tau,\, \tau,\, \EncImp(\mathbf{x}_{\mathrm{obs}})\right)
      - (x_1 - x_0) \right\|^2_{\mathbf{M}}
    \right], \label{eq:loss_fm}
\end{align}
where $x_\tau = (1-\tau)\,x_0 + \tau\,x_1$ is the linear interpolation at FM time
$\tau \in [0,1]$ (distinct from the time-series index $t \in \{1,\ldots,T\}$),
$\mathbf{M} \in \{0,1\}^{T \times C}$ is the binary imputation target mask
(union of all missing positions; not the modality count $M$),
$x_1$ is the ground-truth signal at those positions, and $x_0$ is the prior-aware
source initialization from Section~\ref{sec:prior}.
$\Lds$ is used in Phases~1 and~3 to learn and then refine predictive representations, whereas $\Lfm$ is used in Phase~2 to train the flow matching imputation module and reconstruct the observation from the prior initialization. 

With these objectives in place, we train the framework in three phases.

\textbf{Phase~1: Downstream task warmup.}
We train $\EncFus$ and $\fds$ with $\Lds$ on the raw, partially
observed multimodal input $\mathbf{x}_{\mathrm{obs}}$, without any
imputed augmentation.
Starting from incomplete data forces the encoder to learn
representations that remain informative under missingness,
rather than relying on completed signals.

\textbf{Phase~2: Flow matching training.}
$\EncImp$ is initialized by copying the weights of $\EncFus$ so
that the imputation module starts from representations already
shaped by the downstream task objective.
The FM learns to impute conditioned on $\mathbf{c}_\mathrm{imp}$,
which encodes cross-modal context from the present modalities and
provides the velocity field with information about the content of
the missing signal.

\textbf{Phase~3: Joint training on completed data.}
After Phase~2, $\EncImp$ and $\ffm$ are frozen and used to impute
all missing positions in producing the completed input
$\mathbf{x}_{\mathrm{complete}}$.
We then train $\EncFus$ and $\fds$ with $\Lds$ on $\mathbf{x}_{\mathrm{complete}}$ to get final result.

Through this mechanism, weight sharing lets the FM branch inherit
task-relevant encoder parameters, while imputed-data fine-tuning allows the
recovered signals to improve the final downstream model; the effectiveness of this
strategy is experimented and shown in Figure~\ref{fig:strategy_ablation}.

\section{Experiments}
\label{sec:experiments}

\subsection{Experimental Setup}
\label{sec:setup}
\paragraph{Datasets.}
We evaluate on four physiological benchmarks. \textbf{Sleep-EDF}~\cite{kemp2009sleep}
is a polysomnography dataset for sleep staging, which we represent with four
modalities: EEG (2 channels), EOG (1 channel), Resp (1 channel), and EMG
(1 channel), and use for five-class sleep stage classification.
\textbf{PTB-XL}~\cite{wagner2020ptb} is a 12-lead ECG benchmark, which we
partition into three lead-group modalities, Limb (I/II/III), Augmented
(aVR/aVL/aVF), and Precordial (V1--V6), for five-class cardiac diagnostic
classification (NORM/MI/HYP/CD/STTC). \textbf{PPG-DaLiA}~\cite{reiss2019deep}
is a wearable sensing dataset originally released for heart-rate estimation;
in our benchmark-adapted setting, we use five sensor modalities, chest ACC,
wrist ACC, wrist BVP, wrist EDA, and wrist TEMP, for nine-class activity
recognition. \textbf{Chapman-Shaoxing}~\cite{zheng202012} is a 12-lead
arrhythmia corpus, which we represent as twelve single-lead modalities and
use for 7-label diagnosis prediction. Detailed dataset statistics, split
protocols, and preprocessing are provided in Appendix~\ref{app:datasets}.

\paragraph{Missing data protocol.}
We evaluate missingness under both basic and stress-test protocols.
As basic settings, we consider \wmissing{} and \mmissing{} separately at a
20\% missing rate.
For \wmissing{}, instead of adding Gaussian noise, we apply contiguous block
masks with lengths drawn uniformly from $[5\%, 10\%]$ of $T$, per channel, to mimic missingness from transient sensor dropout
or signal failure.
For \mmissing{}, we remove whole modalities to mimic missing from no recoding, for example, subjects who did not wear or record a given sensor. 
Beyond these basic settings, we further evaluate a mixed protocol where
\wmissing{} and \mmissing{} occur simultaneously to simulate a more realistic clinical setting, for example, in ICU a patient may lack EEG record entirely while recorded vital signs are also
temporally interrupted~\cite{johnson2020mimic}, as well as high-missingness
protocols that increase the missing rate to test robustness under more severe
observation loss.

\paragraph{Baselines and Metrics.}
We compare against four strong multimodal methods that handle missing modalities within two years.
FuseMoE~\cite{han2024fusemoe} routes inputs through modality-specific experts,
tolerating absent streams via sparse gating.
Flex-MoE~\cite{yun2024flexmoe} substitutes missing modalities with learned
embeddings in a mixture-of-experts architecture.
Maestro~\cite{mohapatra2025maestro} applies adaptive sparse attention over partially observed modality streams.
MIRA~\cite{li2025mira} is a channel-independent sparse MoE Transformer
trained for multimodal classification under missingness.
None of these methods explicitly distinguishes between \wmissing{} and \mmissing{},
nor do they reconstruct missing signal values.
We report AUROC and macro-averaged F1 score for classification,
averaged over available independent runs with 3 random seeds.

\subsection{Comparison with Baselines}
\label{sec:main_results}


\begin{table*}[t]
\centering
\caption{Classification performance on four benchmarks under 20\% missingness,
  with \wmissing{} and \mmissing{} evaluated separately.
  Means and standard deviations over three runs are rounded to three decimals.
  \textbf{Bold}: best; \underline{underlined}: second-best.}
\label{tab:main}
\vspace{.5em}
\setlength{\tabcolsep}{0.5pt}
\scriptsize
\begin{tabular}{l cc cc cc cc}
\toprule[1pt]
& \multicolumn{2}{c}{\textbf{Sleep-EDF}}
& \multicolumn{2}{c}{\textbf{PTB-XL}}
& \multicolumn{2}{c}{\textbf{PPG-DaLiA}}
& \multicolumn{2}{c}{\textbf{Chapman}} \\
\cmidrule(lr){2-3}\cmidrule(lr){4-5}\cmidrule(lr){6-7}\cmidrule(lr){8-9}
\textbf{Method} & F1 & AUROC & F1 & AUROC & F1 & AUROC & F1 & AUROC \\
\midrule
\multicolumn{9}{l}{\textit{\wmissing{} (20\%)}} \\[1pt]
FuseMoE~\cite{han2024fusemoe}       & $0.552_{\pm 0.007}$ & $0.912_{\pm 0.006}$ & $0.243_{\pm 0.020}$ & $0.683_{\pm 0.011}$ & $0.602_{\pm 0.014}$ & $0.918_{\pm 0.008}$ & $0.311_{\pm 0.023}$ & $0.630_{\pm 0.007}$ \\
Flex-MoE~\cite{yun2024flexmoe}      & $\underline{0.668}_{\pm 0.007}$ & $\underline{0.967}_{\pm 0.003}$ & $0.422_{\pm 0.022}$ & $0.731_{\pm 0.009}$ & $0.591_{\pm 0.005}$ & $0.917_{\pm 0.004}$ & $0.343_{\pm 0.003}$ & $0.708_{\pm 0.008}$ \\
Maestro~\cite{mohapatra2025maestro} & $0.627_{\pm 0.003}$ & $0.955_{\pm 0.000}$ & $\underline{0.622}_{\pm 0.007}$ & $\underline{0.876}_{\pm 0.001}$ & $\underline{0.692}_{\pm 0.017}$ & $\underline{0.943}_{\pm 0.001}$ & $0.351_{\pm 0.024}$ & $0.762_{\pm 0.009}$ \\
MIRA~\cite{li2025mira}              & $0.582_{\pm 0.013}$ & $0.936_{\pm 0.010}$ & $0.531_{\pm 0.004}$ & $0.839_{\pm 0.008}$ & $0.672_{\pm 0.002}$ & $0.936_{\pm 0.002}$ & $\underline{0.403}_{\pm 0.013}$ & $\underline{0.815}_{\pm 0.017}$ \\
\method{} (ours)                    & $\mathbf{0.697}_{\pm 0.023}$ & $\mathbf{0.977}_{\pm 0.001}$ & $\mathbf{0.661}_{\pm 0.003}$ & $\mathbf{0.884}_{\pm 0.003}$ & $\mathbf{0.714}_{\pm 0.017}$ & $\mathbf{0.948}_{\pm 0.006}$ & $\mathbf{0.518}_{\pm 0.037}$ & $\mathbf{0.870}_{\pm 0.007}$ \\
\midrule
\multicolumn{9}{l}{\textit{\mmissing{} (20\%)}} \\[1pt]
FuseMoE~\cite{han2024fusemoe}       & $0.502_{\pm 0.005}$ & $0.890_{\pm 0.001}$ & $0.314_{\pm 0.019}$ & $0.662_{\pm 0.004}$ & $0.562_{\pm 0.005}$ & $0.901_{\pm 0.004}$ & $0.327_{\pm 0.003}$ & $0.644_{\pm 0.002}$ \\
Flex-MoE~\cite{yun2024flexmoe}      & $\underline{0.651}_{\pm 0.005}$ & $\underline{0.961}_{\pm 0.001}$ & $0.447_{\pm 0.029}$ & $0.728_{\pm 0.009}$ & $0.570_{\pm 0.007}$ & $0.909_{\pm 0.000}$ & $0.357_{\pm 0.010}$ & $0.699_{\pm 0.006}$ \\
Maestro~\cite{mohapatra2025maestro} & $0.646_{\pm 0.012}$ & $0.953_{\pm 0.003}$ & $\underline{0.604}_{\pm 0.010}$ & $\underline{0.872}_{\pm 0.002}$ & $\underline{0.675}_{\pm 0.013}$ & $0.931_{\pm 0.003}$ & $0.413_{\pm 0.015}$ & $0.758_{\pm 0.003}$ \\
MIRA~\cite{li2025mira}              & $0.561_{\pm 0.003}$ & $0.925_{\pm 0.005}$ & $0.504_{\pm 0.007}$ & $0.826_{\pm 0.001}$ & $0.663_{\pm 0.003}$ & $\underline{0.932}_{\pm 0.003}$ & $\underline{0.415}_{\pm 0.024}$ & $\mathbf{0.827}_{\pm 0.010}$ \\
\method{} (ours)                    & $\mathbf{0.653}_{\pm 0.009}$ & $\mathbf{0.963}_{\pm 0.005}$ & $\mathbf{0.648}_{\pm 0.002}$ & $\mathbf{0.878}_{\pm 0.002}$ & $\mathbf{0.691}_{\pm 0.015}$ & $\mathbf{0.944}_{\pm 0.005}$ & $\mathbf{0.492}_{\pm 0.012}$ & $\underline{0.823}_{\pm 0.006}$ \\
\bottomrule[1pt]
\vspace{-2em}
\end{tabular}
\end{table*}

\paragraph{Basic missingness.}
Table~\ref{tab:main} reports classification performance under the two basic
20\% missingness settings. Since all four benchmarks involve skewed label
distributions across sleep stages, diagnosis classes, or activity labels, and
precision-recall-based evaluation is often more informative than ROC-based
evaluation on imbalanced classification problems~\cite{davis2006relationship, saito2015precision},
we treat macro-F1 as the primary metric in this clinical-style setting.
Three observations are most important. (1) \method{} delivers the strongest
overall classification performance, achieving the best F1 on all eight
dataset-setting pairs and the best AUROC on seven of them, which supports the
advantage of explicit imputation over only tolerating missing inputs. (2) The
strongest baseline differs across datasets, yet our method still performs best overall: Flex-MoE is especially competitive
on Sleep-EDF, particularly under \mmissing{}, suggesting that its
subset-aware routing and missing-modality bank are effective when whole
modalities disappear; Maestro's modality-aware encoding helps it to become the strongest baseline on PTB-XL and PPG-DaLiA; and MIRA is strongest on Chapman, suggesting that its continuous-time
sparse modeling is helpful on this more challenging ECG benchmark. Even so,
\method{} remains best overall, indicating that explicit imputation provides a
more consistent route to downstream accuracy. (3) The clearest advantage
appears in macro-F1, which is especially relevant here because minority classes
matter in all four benchmarks and F1 is often more informative than AUROC for
imbalanced clinical-style classification~\cite{davis2006relationship, saito2015precision}.

\paragraph{Mixed and high missingness.}
Beyond the two basic 20\% settings, Table~\ref{tab:mixed_high} evaluates
harder but realistic scenarios. The mixed setting combines \wmissing{} and
\mmissing{} in the same sample, which better reflects clinical data. We also consider a more extreme 50\%
missingness setting to test robustness under severe observation loss. As Table~\ref{tab:mixed_high} shows, \method{} maintains the strongest overall downstream performance across all
three settings. Under mixed missingness, it achieves the best F1 and matches
the best AUROC, showing that its advantage persists when the two missingness
mechanisms co-occur. Under 50\% missingness, it continues to rank first on
both settings. The clearest contrast is with
Maestro under \wmissing{}, where F1 falls from 0.622 in the 20\% basic setting
to 0.514 at 50\%, a drop of 0.108. It is not surprising that while
Maestro is strong with modality-level missing awareness, it does not specifically model
within-modality gaps, so performance degrades more sharply when \wmissing{} becomes heavy. Compared with Maestro, our imputation strategy is more resilient under heavy missingness.

\begin{table*}[t]
\centering
\caption{PTB-XL classification under mixed and high missingness.
  Mixed denotes simultaneous \wmissing{}+\mmissing{} at 20\% each.
  High-rate settings evaluate either \wmissing{} or \mmissing{} at 50\%.
  Means and standard deviations over three runs are rounded to three decimals.
  \textbf{Bold}: best; \underline{underlined}: second-best.}
\label{tab:mixed_high}
\vspace{.5em}
\setlength{\tabcolsep}{7pt}
\scriptsize
\begin{tabular}{l cc cc cc}
\toprule[1pt]
& \multicolumn{2}{c}{\textbf{Mixed}}
& \multicolumn{2}{c}{\textbf{\wmissing{} 50\%}}
& \multicolumn{2}{c}{\textbf{\mmissing{} 50\%}} \\
\cmidrule(lr){2-3}\cmidrule(lr){4-5}\cmidrule(lr){6-7}
\textbf{Method} & F1 & AUROC & F1 & AUROC & F1 & AUROC \\
\midrule
FuseMoE~\cite{han2024fusemoe}
& $0.211_{\pm 0.040}$ & $0.601_{\pm 0.003}$
& $0.176_{\pm 0.041}$ & $0.612_{\pm 0.002}$
& $0.237_{\pm 0.014}$ & $0.617_{\pm 0.021}$ \\
Flex-MoE~\cite{yun2024flexmoe}
& $0.428_{\pm  0.024}$ & $0.757_{\pm  0.003}$
& $0.411_{\pm 0.008}$ & $0.708_{\pm 0.006}$
& $0.460_{\pm 0.005}$ & $0.709_{\pm 0.006}$ \\
Maestro~\cite{mohapatra2025maestro}
& $\underline{0.608}_{\pm 0.009}$ & $\mathbf{0.867}_{\pm 0.002}$
& $\underline{0.514}_{\pm 0.012}$ & $\underline{0.856}_{\pm 0.003}$
& $\underline{0.566}_{\pm 0.030}$ & $\underline{0.848}_{\pm 0.004}$ \\
MIRA~\cite{li2025mira}
& $0.503_{\pm 0.005}$ & $0.828_{\pm 0.003}$
& $0.455_{\pm 0.004}$ & $0.812_{\pm 0.001}$
& $0.505_{\pm 0.001}$ & $0.823_{\pm 0.005}$ \\
\method{} (ours)
& $\mathbf{0.641}_{\pm 0.011}$ & $\mathbf{0.867}_{\pm 0.003}$
& $\mathbf{0.626}_{\pm 0.011}$ & $\mathbf{0.865}_{\pm 0.003}$
& $\mathbf{0.600}_{\pm 0.012}$ & $\mathbf{0.852}_{\pm 0.005}$ \\
\bottomrule[1pt]

\end{tabular}
\end{table*}

\subsection{Ablation Study}
\label{sec:ablation}
\paragraph{Hyperparameter.}
\begin{figure*}[t]
  \centering
  \begin{subfigure}[t]{0.48\linewidth}
    \centering
    \includegraphics[width=\linewidth]{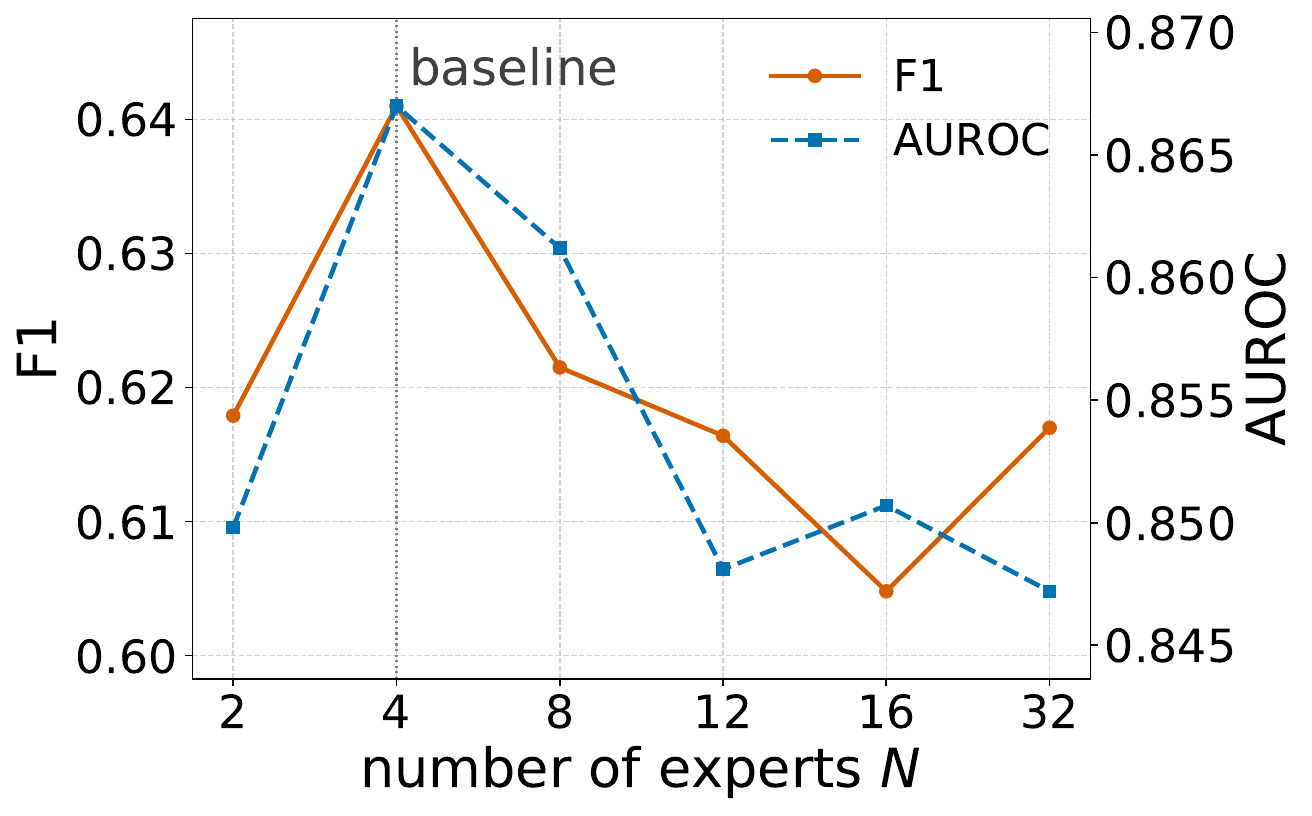}
  \end{subfigure}
  \hfill
  \begin{subfigure}[t]{0.48\linewidth}
    \centering
    \includegraphics[width=\linewidth]{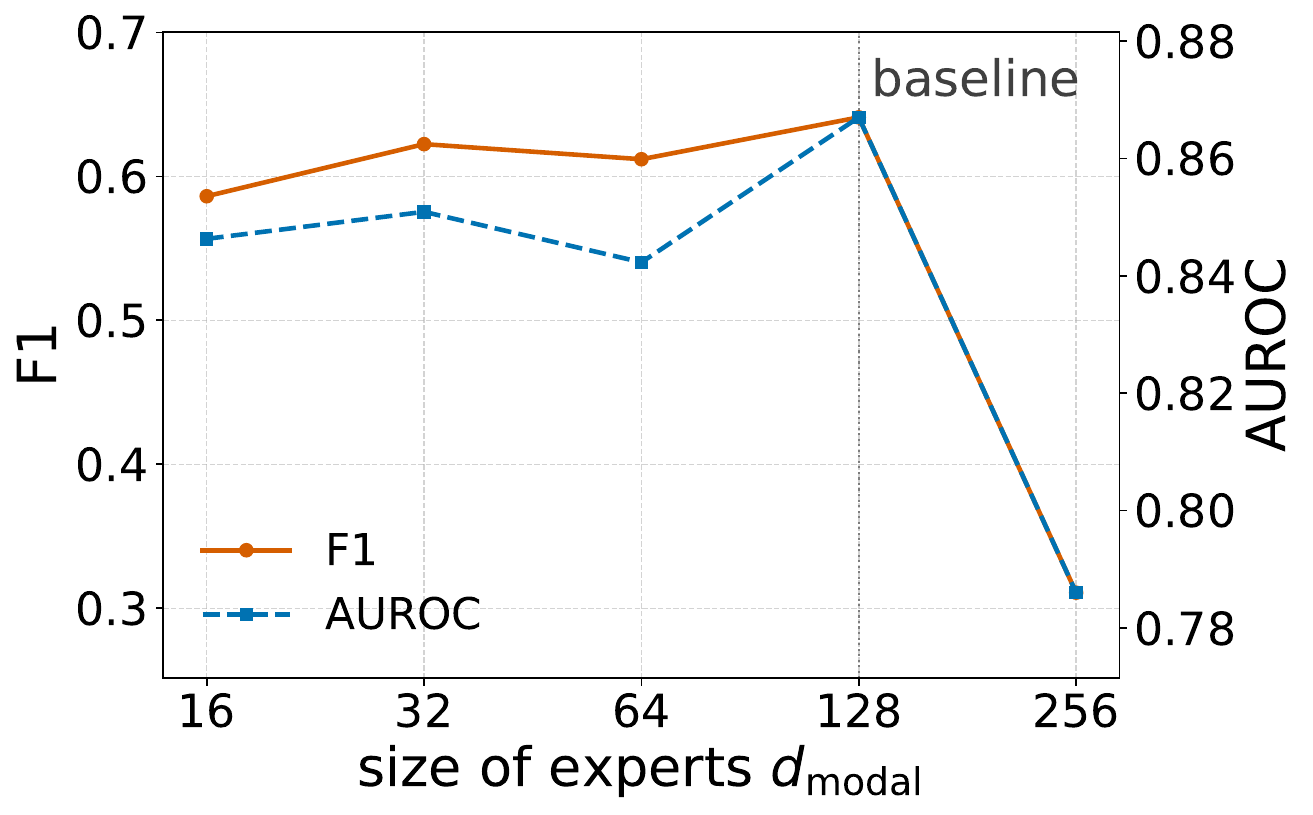}
  \end{subfigure}
  \caption{Hyperparameter sensitivity of our encoder configuration on
  PTB-XL under mixed 20\% missingness. Left: sweep over the number of experts
  $N$. Right: sweep over the expert size $d_{\mathrm{modal}}$.}
  \label{fig:hpo_curves}
\end{figure*}
Our encoder introduces two key capacity controls: the number of
experts $N$ and the expert size $d_{\mathrm{modal}}$. The former determines
how finely the model partitions modality-specific patterns, while the latter
controls the representation capacity of each expert. Both are important for
performance, but both can also become harmful when enlarged without control.
If $N$ is too large, expert specialization becomes overly fragmented, making
routing less stable and reducing effective data per expert. If
$d_{\mathrm{modal}}$ is too large, the expert MLPs become more prone to
overfitting under the limited supervision strategy. We therefore
conduct the hyperparameter sweep in Figure~\ref{fig:hpo_curves} under mixed
20\% missingness. As shown in the left panel, performance is strongest around
$N{=}4$; after that point, the results begin to oscillate and decrease rather than improve,
indicating that adding more experts mainly increases fragmentation rather than
useful specialization. In the right panel, performance increases gradually
before reaching its best point at $d_{\mathrm{modal}}{=}128$, after which it
drops sharply. Based on this trend, we use $N{=}4$ and
$d_{\mathrm{modal}}{=}128$ in the final configuration.

\begin{wrapfigure}{r}{0.44\linewidth}
  \vspace{-1.2em}
  \centering
  \includegraphics[width=\linewidth]{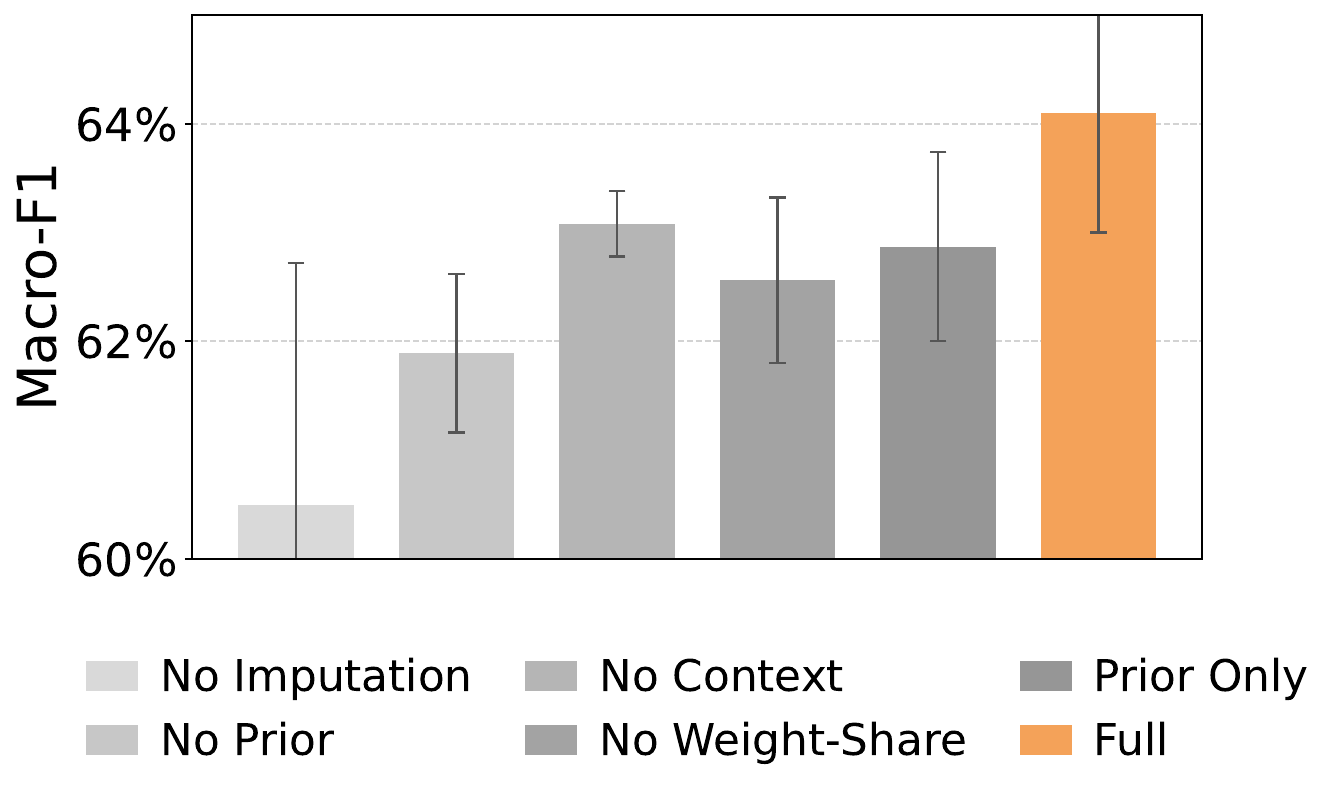}
  \caption{Strategy ablation on PTB-XL under mixed 20\% missingness. All variants underperform the full model.}
  \label{fig:strategy_ablation}
  \vspace{-1em}
\end{wrapfigure}

Figure~\ref{fig:strategy_ablation} summarizes the effect of removing each major
design component under mixed 20\% missingness. All variants
underperform the full model, confirming that the final performance relies on
the cooperation of imputation, prior-aware source design, conditional
cross-modal guidance, and encoder-level knowledge transfer. Among them,
``No Imputation'' yields the worst result, showing that directly optimizing the
classifier on raw incomplete inputs is insufficient in the missing
setting. Removing prior-aware initialization also leads to a clear performance
drop and larger fluctuations, suggesting that a structured source
initialization is important for stabilizing the conditional generation process.
Similarly, ``No Context'' degrades performance,
indicating that conditioning the generative pathway on encoder-side
context is also beneficial. ``No Weight-Share''
produces consistently lower results. 
We believe that without weight sharing, the imputation branch no longer
receives sufficient guidance from the shared encoder representation shaped by
the downstream classification objective. The ``Prior Only'' control performs
better than ``No Imputation'' but still remains below the full model,
indicating that the complete pipeline benefits from more than the prior signal
alone.

\subsection{Comparison with Diffusion Imputation}
\label{sec:efficiency}
\paragraph{Strategy Effectiveness.}
We compare \method{} against 
diffusion-based imputation methods to assess
whether flow matching provides a stronger generative imputation module under
the same downstream task setting. For a fair comparison, we pair each
diffusion baseline with the same downstream task component used in our
framework, so the comparison isolates the generative imputation module rather
than conflating it with differences in the classifier. Specifically, we
evaluate the performance of CSDI~\cite{tashiro2021csdi} and SSSD~\cite{alcaraz2023sssd} on
PTB-XL under mixed 20\% missingness. Table~\ref{tab:efficiency} shows that
\method{} achieves the highest Macro-F1 while also
delivering the lowest inference latency. This speedup comes from the FM
transport, which reaches $x_1$ from $x_0$ with fewer function evaluations than
iterative diffusion denoising. The performance gain is also consistent with
our design: The source stage $x_0$ is
initialized by the prior-aware rule in Equation~\eqref{eq:prior_aware}, so the
transport starts from a more informative point than the generic diffusion
initialization used by CSDI and SSSD. Moreover, because these baselines are primarily designed for per-modality recovery, they are less well suited to
modality-level missingness, where imputation must rely on cross-modal evidence rather than within-modality continuity.

\begin{table}[t]
\centering
\caption{Comparison with diffusion-based imputation methods on PTB-XL under mixed 20\% missingness (batch size = 64, GPU: B200).
  Each method is paired with the same downstream task component to isolate the generative imputation module.
  Performance is reported as Macro-F1.
  NFE denotes the number of function evaluations, and latency is wall-clock time per batch in milliseconds.
}
\label{tab:efficiency}
\vspace{.5em}
\small
\setlength{\tabcolsep}{25pt}
\begin{tabular}{lccc}
\toprule[1pt]
\textbf{Method} & \textbf{Macro-F1} & \textbf{NFE} & \textbf{Latency (ms/batch)} \\
\midrule 
CSDI~\cite{tashiro2021csdi}         & $0.626_{\pm 0.016}$ & 50 & 137.77 \\
SSSD~\cite{alcaraz2023sssd}         & $0.616_{\pm 0.009}$ & 50 & 526.98 \\
\method{} (ours)                           & $\textbf{0.641}_{\pm 0.011}$ & \textbf{20} & \textbf{75.07} \\
\bottomrule[1pt]
\vspace{-2em}
\end{tabular}
\end{table}


\section{Conclusion}
\label{sec:conclusion}

Missing data remains a major obstacle in multimodal time series analysis, because both \wmissing{} and \mmissing{} can severely degrade downstream performance. \method{} addresses this gap through prior-aware initialization in a flow-matching framework and through weight sharing with imputed-data fine-tuning that connects imputation to classification. Experiments on four datasets show that this combination delivers the strongest overall downstream performance against missing-aware baselines and direct imputation pipelines, supporting the value of distinguishing missingness structure and coupling signal recovery with downstream learning.

\bibliographystyle{plainnat}
\bibliography{references}

\appendix

\section{Limitations}
\label{app:limitations}

Although \method{} improves both imputation and downstream prediction under
diverse missingness settings, it still has several limitations. First, because
the imputation branch is learned rather than analytically specified, the method
still requires partially observed multimodal training data rich enough to learn
reliable temporal and cross-modal structure. Second, our empirical study covers four public
healthcare benchmarks under controlled missingness protocols, so the observed
performance may depend on the benchmark preprocessing, the chosen missingness
simulation, and the strength of cross-modal correlations available in each
dataset.

\section{Broader Impacts}
\label{app:broader}

This work studies missing-aware multimodal time-series modeling for healthcare,
so its main potential positive impact is improved robustness when some sensors
or clinical measurements are unavailable. In settings where full multimodal
acquisition is difficult, a model that better handles both \wmissing{} and
\mmissing{} may help maintain downstream predictive performance and make
multimodal decision support more usable in practice.

At the same time, the method also carries several risks. First, explicit
imputation can produce plausible but incorrect physiological signals, which may
encourage over-trust if the completed inputs are treated as ground truth rather
than model estimates. Second, because the approach depends on learned temporal
and cross-modal correlations, performance may degrade under distribution shift,
different missingness mechanisms, or populations that are underrepresented in
the training data, which could amplify unfair error patterns in deployment.
Third, the target application domain involves sensitive health data, so any
real-world use would require privacy-preserving data governance, secure access
control, and clinical oversight. For these reasons, we view \method{} as a
research-stage method rather than a deployment-ready clinical system, and any
practical use should include site-specific validation and monitoring under
missingness.

\section{Prior-Aware Source Initialization}
\label{app:prior}

This appendix provides the complete initialization formulas for the
prior-aware source initialization $x_0$ introduced in
Section~\ref{sec:prior}, along with the design-choice ablation that
justifies the configuration used in our main experiments.

\subsection{Within-Modality Prior}

For each missing position $(t, c) \in \Omega_w$, we initialize
$x_0^{\mathrm{wm}}[t,c]$ as a convex combination of the nearest observed
temporal neighbor and the per-channel batch mean:
\begin{equation}
  x_0^{\mathrm{wm}}[t,c]
  = \alpha_{t,c}\, x^{\mathrm{nbr}}_{t,c}
  + (1 - \alpha_{t,c})\, \bar{x}_c^{\mathrm{batch}},
  \label{eq:prior_wm}
\end{equation}
where $x^{\mathrm{nbr}}_{t,c}$ is the value of the nearest observed entry
in the same channel:
\begin{equation}
  x^{\mathrm{nbr}}_{t,c}
  = x_{t^*, c}, \quad
  t^* = \arg\min_{\,t' :\, R^{(m)}_{t',c}=1}\, |t' - t|,
  \label{eq:nbr}
\end{equation}
Here $R^{(m)}$ is the observation mask from Section~\ref{sec:problem}, and
$\bar{x}_c^{\mathrm{batch}}$ is the fallback batch mean when no neighbor exists
in channel $c$.

The blending weight $\alpha_{t,c}$ encodes confidence in the neighbor: it
should be high when the gap is short and decay as the gap widens.
We consider two instantiations:

\paragraph{Rule-based (deterministic).}
\begin{equation}
  \alpha_{t,c} = \exp\!\left(-\lambda \cdot d_{t,c}\right),
  \quad
  d_{t,c} = |t^* - t|,
  \label{eq:alpha_rule}
\end{equation}
where $\lambda > 0$ is a fixed decay coefficient
(default $\lambda = 0.1$; ablated in Table~\ref{tab:prior_ablation}).
When no observed neighbor exists, $\alpha_{t,c} = 0$ (pure batch mean).

\paragraph{Learned (adaptive).}
\begin{equation}
  \alpha_{t,c} = \sigma\!\left(\mathrm{MLP}(d_{t,c},\, \mathbf{c}_{\mathrm{imp}})\right),
  \label{eq:alpha_learned}
\end{equation}
where $\sigma$ is the sigmoid function and $\mathbf{c}_{\mathrm{imp}}$
provides cross-modal context, allowing the network to adapt the
confidence based on the broader signal structure.

The rule-based variant requires no additional parameters and is
theoretically grounded: as noted by~\citet{kollovieh2025tsflow},
temporal neighbor interpolation corresponds to the posterior mean of a
Gaussian process with a locally stationary kernel in the zero-noise
limit, making it a principled approximation that avoids the $O(T^3)$
cost of full GP inference.

\subsection{Modality-Level Prior}

For modality-level missing positions $(t,c) \in \Omega_m$, no
within-modality signal exists.
We derive $x_0^{\mathrm{ml}}$ from the cross-modal context
$\mathbf{c}_{\mathrm{imp}}$ produced by $\EncImp$.
We consider three strategies of increasing complexity:

\paragraph{Batch mean (B1).}
\begin{equation}
  x_0^{\mathrm{ml}} = \bar{x}^{\mathrm{batch}},
  \label{eq:prior_ml_mean}
\end{equation}
where $\bar{x}^{\mathrm{batch}} \in \mathbb{R}^{T \times C_m}$ is the
per-channel mean over all samples in the current batch that have modality
$m$ observed.
This strategy is simple and parameter-free but ignores sample-level
heterogeneity.

\paragraph{Batch matching (B2).}
For a query sample $b$ with missing modality $m$, we retrieve the most
context-similar sample in the batch that has modality $m$ observed:
\begin{equation}
  b^* = \arg\max_{b' \in \mathcal{B},\, m \in \mathrm{obs}(b')}
    \cos\!\left(\mathbf{c}_{\mathrm{imp}}^{(b)},\,
    \mathbf{c}_{\mathrm{imp}}^{(b')}\right),
  \quad
  x_0^{\mathrm{ml}} = x^{(b^*)}_{m},
  \label{eq:prior_ml_match}
\end{equation}
Here $\mathcal{B}$ is the current batch, $\mathrm{obs}(b')$ denotes
the observed modalities of sample $b'$, and $x_m^{(b^*)}$ is modality $m$ from
the matched sample $b^*$.
The most context-similar sample that has modality $m$ present provides
the initialization; if no such sample exists in the batch, the method
falls back to B1.

\paragraph{Memory bank (B3).}
A set of $P$ learnable prototype vectors
$\mathcal{C}_m = \{c_p\}_{p=1}^P \subset \mathbb{R}^{T \times C_m}$
is maintained per modality.
The initialization is a soft retrieval:
\begin{equation}
  x_0^{\mathrm{ml}}
  = \sum_{p=1}^{P} \pi_p\, c_p,
  \quad
  \pi_p = \frac{\exp\!\left(\cos\!\left(\mathbf{c}_{\mathrm{imp}},\,
    \tilde{c}_p\right)\right)}
    {\sum_{p'} \exp\!\left(\cos\!\left(\mathbf{c}_{\mathrm{imp}},\,
    \tilde{c}_{p'}\right)\right)},
  \label{eq:prior_ml_bank}
\end{equation}
where $\tilde{c}_p$ is the key associated with prototype $c_p$, and $\pi_p$ is
its softmax retrieval weight.
This variant generalizes across batches but introduces $P$ additional
parameters per modality.

These three modality-level priors reflect a clear trade-off between simplicity,
sample adaptivity, and parameterization. The batch-mean baseline is maximally
stable and easy to compute, but it ignores instance-specific variation. Batch
matching injects sample-specific information through cross-modal similarity,
yet its quality depends on whether the current mini-batch contains a suitable
reference example. The memory-bank design tries to remove that batch dependence
by storing reusable prototypes, but doing so also compresses the space of
possible initializations into a small learned dictionary. To make these
trade-offs explicit, we report a quantitative comparison of the prior variants
later in Appendix~\ref{app:additional}.

\section{Dataset Details}
\label{app:datasets}

\subsection{Dataset Overview}
\label{app:dataset_overview}

The four benchmarks used in Section~\ref{sec:setup} cover sleep staging, ECG
diagnosis, wearable activity recognition, and rhythm classification, for a
total of 108,098 processed samples. Although all four datasets are
physiological time series, they stress the benchmark in different ways:
modality counts range from 3 grouped lead streams to 12 single-lead streams,
channel counts range from 5 to 12, and sequence lengths range from 250 to
1000 time steps after preprocessing. This diversity is useful for our
missingness study because it tests whether the same prior-aware and
missingness-aware design can remain effective across both compact and
fine-grained multimodal layouts.

To keep the appendix readable, each dataset summary below follows the same
structure: source and task, modality construction, and preprocessing with split
policy. This standardization makes it easier to compare benchmark assumptions
across datasets without overloading the main paper's experiment section.
At the same time, the four datasets differ not only in scale but also in how
the multimodal structure is defined. Sleep-EDF and PPG-DaLiA combine signals
from different physiological sources, whereas PTB-XL and Chapman-Shaoxing use
ECG-based modality decompositions that preserve clinically meaningful lead
organization. As a result, the benchmark covers both settings where modalities
are semantically distinct by nature and settings where the multimodal view is
constructed from a larger sensing system to expose structured missingness.

Another important source of variation is temporal resolution. Sleep-EDF starts
from long 30-second physiological epochs, PTB-XL and Chapman-Shaoxing come
from fixed-length diagnostic ECG recordings, and PPG-DaLiA is built from
shorter sliding windows over free-living wearable streams. These differences
matter for our study because within-modality missingness and modality-level
missingness can interact differently with long sequences, compact diagnostic
snapshots, and motion-heavy activity windows.

\subsection{Per-Dataset Details}
\label{app:dataset_profiles}

\paragraph{Sleep-EDF dataset.}
The Sleep-EDF dataset provides overnight polysomnographic recordings with
manual hypnograms and standard sleep-related physiological
signals~\cite{kemp2009sleep}. In our benchmark, it serves as a
heterogeneous-but-aligned multimodal sleep-staging task, where all modalities
share the same 30-second epoch but contribute different physiological evidence.

\begin{itemize}
  \item \textbf{Task.} Five-class sleep-stage classification.
  \item \textbf{Labels.} `W` = wakefulness; `N1` = light sleep onset; `N2` = stable light sleep with spindles/K-complexes; `N3` = deep slow-wave sleep (with S3/S4 merged); `REM` = rapid-eye-movement sleep.
  \item \textbf{Modalities.} We retain five channels and group them into four modalities: EEG (2 channels), EOG (1 channel), Resp (1 channel), and EMG (1 channel). This preserves clinically meaningful separation between neural, ocular, respiratory, and muscle-tone signals. Body-temperature channels available in some releases are not used.
  \item \textbf{Preprocessing and splits.} The loader uses fixed 30-second epochs from the benchmark-provided \texttt{split\_v2.json} partition, yielding 10,918 processed samples. Each epoch is downsampled to 250 time steps and normalized with per-sample z-score scaling at load time, producing inputs of shape $\mathbb{R}^{5 \times 250}$.
\end{itemize}

\paragraph{PTB-XL dataset.}
PTB-XL is a large public 12-lead ECG benchmark with coarse diagnostic
annotations over standard clinical superclasses~\cite{wagner2020ptb}. In our
benchmark, it represents a structured ECG setting where the multimodal view is
constructed from clinically related lead groups rather than distinct sensor
families.

\begin{itemize}
  \item \textbf{Task.} Multi-label ECG diagnosis prediction with five diagnostic superclasses.
  \item \textbf{Labels.} `NORM` = normal ECG; `MI` = myocardial infarction; `STTC` = ST/T-wave change; `CD` = conduction disturbance; `HYP` = hypertrophy. A record may activate more than one label.
  \item \textbf{Modalities.} We split the twelve leads into three lead-group modalities: Limb (I/II/III), Augmented (aVR/aVL/aVF), and Precordial (V1--V6). This grouping gives modality-level missingness a concrete interpretation as the loss of an anatomically related lead group rather than a single isolated channel.
  \item \textbf{Preprocessing and splits.} The repository loader uses split files keyed by \texttt{ecg\_id}. After preprocessing, each sample is represented with 12 channels and 250 time steps, i.e., $\mathbb{R}^{12 \times 250}$. Per-lead normalization is applied before classification and missingness evaluation.
\end{itemize}

\paragraph{PPG-DaLiA dataset.}
PPG-DaLiA is a wearable sensing dataset collected from 15 subjects in
free-living activity settings, with synchronized chest and wrist recordings and
ECG-based heart-rate reference signals~\cite{reiss2019deep}. In our benchmark,
we use it as a multimodal wearable activity-recognition task, which makes it a
natural testbed for missingness under motion-heavy, sensor-diverse conditions.

\begin{itemize}
  \item \textbf{Task.} Nine-class activity classification.
  \item \textbf{Labels.} `transient` = transition windows between stable activities; `sitting` = seated rest; `ascending\_stairs` = walking upward on stairs; `descending\_stairs` = walking downward on stairs; `table\_soccer` = playing table soccer; `cycling` = cycling activity; `driving\_car` = driving; `lunch\_break` = eating or quiet break period; `walking` = ordinary walking.
  \item \textbf{Modalities.} We use five sensor modalities defined by the current code path: chest ACC (3 axes), wrist ACC (3 axes), wrist BVP, wrist EDA, and wrist TEMP. This gives 9 total channels and preserves the distinction between motion-dominant and physiology-dominant wearable measurements rather than merging them into a single generic stream.
  \item \textbf{Preprocessing and splits.} The benchmark loader operates on segmented 8-second windows mapped to a common 256-step grid, uses benchmark split JSONs, and produces 64,697 processed segments in total. Per-channel z-score normalization is applied within each segment, giving final inputs of shape $\mathbb{R}^{9 \times 256}$.
\end{itemize}

\paragraph{Chapman-Shaoxing dataset.}
Chapman-Shaoxing is a 12-lead ECG corpus with diagnostic annotations covering
common and additional cardiovascular conditions~\cite{zheng202012}. In our
benchmark-aligned configuration, it complements PTB-XL by exposing a more
fine-grained lead-wise multimodal formulation, where each ECG lead is treated
as its own modality.

\begin{itemize}
  \item \textbf{Task.} Multi-label diagnosis prediction with seven superclasses.
  \item \textbf{Labels.} `normal` = normal rhythm or morphology; `cd` = conduction disturbance; `mi` = myocardial infarction; `sttc` = ST/T-wave change; `other` = other cardiac abnormality not covered by the previous groups; `afib` = atrial fibrillation; `hyp` = hypertrophy. As in PTB-XL, a record may activate multiple labels.
  \item \textbf{Modalities.} We treat each lead as its own modality: I, II, III, aVR, aVL, aVF, and V1--V6. This yields 12 single-lead modalities, so modality-level missingness corresponds to the loss of an individual lead rather than a coarse group.
  \item \textbf{Preprocessing and splits.} The loader resamples each 10-second record from 500\,Hz to 100\,Hz, keeps 1000 time steps per lead, applies per-channel z-score normalization, and reads the benchmark \texttt{climb\_splits} for train/val/test evaluation. If those split files are regenerated from the CSV release, the code maps the CSV \texttt{val} partition to test and repartitions the CSV \texttt{train} subset into train/val with a 90/10 split and seed 42. The resulting tensor shape is $\mathbb{R}^{12 \times 1000}$.
\end{itemize}

\subsection{Overall Summary Table}
\label{app:dataset_summary}

Table~\ref{tab:dataset_summary} consolidates the benchmark properties used
throughout our experiments. It is intended as a quick reference for dataset
scale, modality design, and processed input size, complementing the longer
descriptions above. In particular, it highlights that the benchmark spans both
coarse and fine-grained modality decompositions, moderate and large sample
counts, and short to relatively long processed sequences, which together define
the scope of the missingness scenarios studied in the paper.

\begin{table*}[t]
\centering
\scriptsize
\setlength{\tabcolsep}{4.5pt}
\caption{Summary of the four benchmark datasets used in this work. Sample
counts and tensor shapes correspond to the processed benchmark inputs used by
our loaders rather than the raw source files.}
\label{tab:dataset_summary}
\vspace{.5em}
\begin{tabular}{l l l c c c c}
\toprule[1pt]
\textbf{Dataset} & \textbf{Domain} & \textbf{Task} & \textbf{Samples} & \textbf{Modalities} & \textbf{Channels} & \textbf{Processed shape} \\
\midrule
Sleep-EDF & Sleep staging & 5-class classification & 10,918 & 4 & 5 & $\mathbb{R}^{5 \times 250}$ \\
PTB-XL & ECG diagnosis & 5-label prediction & 21,837 & 3 & 12 & $\mathbb{R}^{12 \times 250}$ \\
PPG-DaLiA & Wearable activity recognition & 9-class classification & 64,697 & 5 & 9 & $\mathbb{R}^{9 \times 256}$ \\
Chapman-Shaoxing & ECG rhythm/diagnosis & 7-label prediction & 10,646 & 12 & 12 & $\mathbb{R}^{12 \times 1000}$ \\
\midrule
\textbf{Total / Range} & \textbf{4 tasks} & --- & \textbf{108,098} & 3--12 & 5--12 & $T=250$--$1000$ \\
\bottomrule[1pt]
\end{tabular}
\end{table*}

\subsection{Missingness Protocol Details}
\label{app:missingness_details}

The main paper reports \wmissing{}, \mmissing{}, mixed missingness, and
high-missingness stress tests. Here we summarize how those conditions are
implemented in the benchmark pipeline. Missingness is injected at data-loading
time rather than written out as a separate copy of every dataset, so different
missingness settings can be evaluated without duplicating the underlying
processed signals on disk. To keep the stochastic patterns reproducible, the
injector constructs a local random generator from the pair
(\texttt{seed}, \texttt{sample\_id}), ensuring that the same sample receives
the same missingness pattern whenever the same configuration is rerun.

\paragraph{Modality-level missingness.}
For modality-level missingness, each currently present modality group is
dropped independently with Bernoulli probability $p_{\mathrm{mod}}$. In the
main experiments, we set $p_{\mathrm{mod}}=0.2$ for the basic
\mmissing{} setting. A safeguard enforces that at least one modality remains:
if all groups would be dropped, one dropped group is restored uniformly at
random. Once a group is removed, its entry in \texttt{mod\_mask} is set to zero
and the corresponding row block in the temporal mask is zeroed as well. The
meaning of one dropped modality therefore depends on the benchmark-specific
grouping: one of four physiological groups in Sleep-EDF, one of three lead
groups in PTB-XL, one of five wearable sensor streams in PPG-DaLiA, or one of
twelve single leads in Chapman-Shaoxing.

\paragraph{Within-modality missingness.}
For within-modality missingness, we use contiguous time-block masking rather
than scattered point deletion. For a sequence of length $T$, each block length
is sampled from the interval $[0.05T,\,0.10T]$, matching the 5--10\% range
described in Section~\ref{sec:setup}. The number of blocks is chosen to reach
approximately 20\% total temporal corruption, and non-overlap is enforced by
rejection sampling with a bounded retry budget. These blocks are applied
independently across time-series rows, so different channels or modality
groups typically lose different temporal windows. This design better mimics
sensor dropouts and short recording interruptions than iid masking would.

\paragraph{Mixed and high missingness settings.}
The mixed setting combines the same two mechanisms: 20\% block-based
\wmissing{} together with 20\% Bernoulli \mmissing{}. In implementation, the
temporal block mask is constructed first and then any fully dropped modalities
overwrite their corresponding temporal rows, so modality removal takes
precedence whenever both types affect the same group. For the high-missingness
stress test, we increase the corruption rate to 50\% while keeping the same
missingness semantics, allowing us to measure whether performance degrades
gracefully when observation loss becomes severe.

\section{Evaluation Metric Details}
\label{app:metrics}

For all classification experiments, we report macro-averaged F1 and
macro-averaged AUROC. Let $N$ be the number of test samples and $K$ the number
of classes. For class $k$, let $TP_k$, $FP_k$, and $FN_k$ denote the numbers of
true positives, false positives, and false negatives, respectively. The
class-wise precision and recall are
\begin{equation}
  P_k = \frac{TP_k}{TP_k + FP_k},
  \qquad
  R_k = \frac{TP_k}{TP_k + FN_k},
\end{equation}
and the class-wise F1 score is
\begin{equation}
  F1_k = \frac{2 P_k R_k}{P_k + R_k}.
\end{equation}
The reported macro-F1 is the unweighted average over classes,
\begin{equation}
  \mathrm{Macro}\text{-}F1 = \frac{1}{K} \sum_{k=1}^{K} F1_k.
\end{equation}

For AUROC, we use a one-vs.-rest formulation and then average across classes.
Let $y_{i,k} \in \{0,1\}$ denote the ground-truth indicator of whether sample
$i$ belongs to class $k$, and let $s_{i,k}$ be the predicted confidence score
for that class. The class-wise AUROC is
\begin{equation}
  \mathrm{AUROC}_k
  =
  \frac{1}{|\mathcal{P}_k|\,|\mathcal{N}_k|}
  \sum_{i \in \mathcal{P}_k}
  \sum_{j \in \mathcal{N}_k}
  \mathbb{I}(s_{i,k} > s_{j,k}),
\end{equation}
where $\mathcal{P}_k = \{i \mid y_{i,k}=1\}$ and
$\mathcal{N}_k = \{i \mid y_{i,k}=0\}$ are the positive and negative sample
sets for class $k$. The reported macro-AUROC is
\begin{equation}
  \mathrm{Macro}\text{-}\mathrm{AUROC}
  = \frac{1}{K} \sum_{k=1}^{K} \mathrm{AUROC}_k.
\end{equation}
This matches our implementation, which computes macro-F1 and macro AUROC with a
one-vs.-rest reduction across classes.

\section{Baseline Methods}
\label{app:baseline_missingness}

This section discusses how the representative baselines used in our experiments
handle missing inputs and, more importantly, what they still leave unresolved.
The core distinction in our paper is between methods that tolerate missingness
at the representation level and methods that explicitly reconstruct missing
signal content before downstream prediction.

\paragraph{Flex-MoE~\cite{yun2024flexmoe}.}
Flex-MoE is one of the clearest examples of a missing-aware fusion model. It
explicitly organizes computation around modality subsets through its routers and
missing-modality bank, so it is built for the combinatorial setting in which
different samples arrive with different observed-modality patterns. Among the
classification baselines, this makes it particularly relevant for
\mmissing{}, because the model does not assume that all modalities are present
at test time. Even so, the substitute for a missing modality remains a learned
subset-aware representation rather than a sample-level reconstruction of the
missing signal itself.
\begin{itemize}
  \item \textbf{What it solves.} It handles arbitrary observed-modality subsets explicitly and learns routing behavior that remains effective when whole modalities are absent.
  \item \textbf{What it does not solve.} It does not reconstruct the missing signal trajectory; the missing-modality bank provides a learned placeholder for fusion, not an imputed waveform or sensor trace tied to the current sample.
  \item \textbf{Why this matters here.} This makes Flex-MoE a strong baseline for missing-aware classification, but not a direct answer to the question of whether explicit signal recovery can provide more downstream value.
\end{itemize}

\paragraph{MIRA~\cite{li2025mira}.}
MIRA is stronger than ordinary fusion baselines in the sense that it natively
handles irregular sampling, sparse observations, and masked values within a
large time-series foundation-model framework. Its continuous-time positional
design and expert specialization make it well matched to temporally sparse
clinical signals, and in our benchmark it is the baseline that comes closest to
treating missingness as part of the time-series learning problem rather than as
an afterthought. Still, MIRA remains prediction-oriented: it uses masks to tell
the encoder which observations are reliable, but it does not build an explicit
sample-specific recovery process for unobserved signals.
\begin{itemize}
  \item \textbf{What it solves.} It can process partially observed time series without heavy dataset-specific redesign and is naturally robust to irregular temporal sparsity.
  \item \textbf{What it does not solve.} It does not explicitly reconstruct missing signal values, and it does not separate \wmissing{} from \mmissing{} through different source priors or recovery paths.
  \item \textbf{Why this matters here.} Our setting distinguishes between local gaps inside an observed modality and the complete absence of a modality, whereas MIRA mainly treats both through masked observation encoding.
\end{itemize}

\paragraph{FuseMoE~\cite{han2024fusemoe}.}
FuseMoE is a cross-modal Transformer with sparse expert routing, so it is
designed to keep multimodal fusion effective even when different streams have
different reliability or sampling structure. In our setting, its main advantage
is that missing modalities can be represented by masked or placeholder tokens
while the observed modalities continue to interact through cross-attention.
This makes it a strong benchmark for missing-aware fusion under
\mmissing{}, especially when the remaining modalities still contain enough
evidence for classification.
\begin{itemize}
  \item \textbf{What it solves.} It learns expert-specialized fusion over the observed subset and can continue prediction without requiring every modality to be present.
  \item \textbf{What it does not solve.} The missing modality is bypassed rather than reconstructed, so the model never produces a sample-specific estimate of the absent stream.
  \item \textbf{Why this matters here.} When the missing modality contains complementary rather than redundant information, missing-aware routing alone may preserve stability but still leave task-relevant evidence unrecovered.
\end{itemize}

\paragraph{Maestro~\cite{mohapatra2025maestro}.}
Maestro is the most explicitly robustness-oriented classifier among the fusion
baselines because it combines sparse attention, modality-aware tokenization,
and curriculum modality dropout. That training strategy exposes the model to
progressively harder missingness patterns and helps it learn to classify from
partial evidence instead of collapsing when an input stream disappears. In our
benchmark, this makes Maestro a particularly strong representative of the
``tolerate missingness'' family.
\begin{itemize}
  \item \textbf{What it solves.} It improves downstream robustness to partial observation through training-time modality dropout and missing-aware token handling.
  \item \textbf{What it does not solve.} Its robustness comes from better classification under absent inputs, not from reconstructing the missing waveform or sensor content itself.
  \item \textbf{Why this matters here.} Maestro tests how far carefully trained missing-aware fusion can go, but it still leaves open whether explicit recovery can add information beyond robust masking-based classification.
\end{itemize}

\paragraph{CSDI~\cite{tashiro2021csdi}.}
CSDI addresses the opposite side of the problem from the fusion baselines above:
it is an explicit imputer. Its role in our benchmark is to provide a principled
diffusion-based reconstruction baseline that fills in missing values before the
downstream classifier is applied. This makes it highly relevant for
\wmissing{}, where the central question is not just whether the classifier can
ignore corrupted segments, but whether the underlying signal can be recovered
well enough to improve the final prediction.
\begin{itemize}
  \item \textbf{What it solves.} It explicitly reconstructs missing values and therefore targets signal recovery directly rather than only masking or bypassing corrupted inputs.
  \item \textbf{What it does not solve.} It is primarily designed for within-series imputation and does not directly address multimodal modality-level absence with different priors for different missingness mechanisms.
  \item \textbf{Why this matters here.} CSDI is a strong test of whether generic diffusion-based imputation helps, but not of whether missingness-type-aware initialization and downstream-coupled training help further.
\end{itemize}

\paragraph{SSSD~\cite{alcaraz2023sssd}.}
SSSD is also an explicit diffusion-style imputer, but with a structured
state-space design intended to model temporal dependencies more efficiently.
Within our benchmark, it serves as a second reconstruction baseline that tests
whether stronger temporal generative modeling alone is sufficient to close the
gap under missingness. Like CSDI, however, its focus is still generic signal
completion rather than multimodal missingness reasoning.
\begin{itemize}
  \item \textbf{What it solves.} It reconstructs missing signal values with a dedicated generative model and therefore provides a direct signal-recovery comparison to our method.
  \item \textbf{What it does not solve.} It does not naturally distinguish \wmissing{} from \mmissing{}, and its imputation stage remains separate from downstream task supervision.
  \item \textbf{Why this matters here.} SSSD helps test whether better generic imputation is enough, but it does not answer whether recovery should be conditioned on missingness type and tied more tightly to the end task.
\end{itemize}

\paragraph{Summary of the gap.}
Taken together, the baselines in our study split into two incomplete families.
Missing-aware fusion models can classify with absent inputs but usually do not
reconstruct the missing signal, whereas explicit imputers can reconstruct
signals but are not built to handle multimodal modality-level absence with
type-specific priors or downstream-coupled training. This is precisely the gap
that motivates \method{} in the main paper. The benchmark is therefore not just
comparing stronger versus weaker classifiers; it is comparing three different
responses to missing data: tolerating it, generically imputing it, or
explicitly reconstructing it with priors matched to the missingness type while
still optimizing for the downstream task.

\begin{table*}[t]
\centering
\small
\setlength{\tabcolsep}{25pt}
\renewcommand{\arraystretch}{1.05}
\caption{Model and training configurations used for \method{}.
The encoder output dimension $\mathrm{fused\_dim}$ equals
$N_{\mathrm{mod}} \times 128$.}
\label{tab:impl_hparams}
\vspace{.5em}
\begin{tabular}{c|c}
\toprule[1pt]
\textbf{Configuration Type} & \textbf{Parameter Name / Value} \\
\midrule
Encoder
& Per-modal projection dimension: 128 \\
& Per-modal Informer layers: 2 \\
& Per-modal attention heads: 2 \\
& Cross-modal MoE blocks with Informer attention: 1 \\
& Cross-modal MoE experts: 4 \\
& Cross-modal expert hidden dimension: 64 \\
& Encoder output dimension ($\mathrm{fused\_dim}$): $N_{\mathrm{mod}} \times 128$ \\
& Classifier dropout: 0.1 \\
\midrule
Flow Matching
& Residual channels: 64 \\
& Number of residual blocks: 4 \\
& FM time embedding dimension: 128 \\
& Feature embedding dimension: 64 \\
& Time-attention heads: 8 \\
& Side-information dimension: $128 + 64 + 1 + \mathrm{fused\_dim}$ \\
& Euler ODE steps: 20 \\
\midrule
Training and Prior
& Phase 1 epochs: 5 \\
& Phase 2 epochs: 20 \\
& Phase 3 epochs: 20 \\
& Batch size: 64 \\
& Classification learning rate: 1e-3 \\
& FM learning rate: 5e-5 \\
& Prior initialization: neighbor\_rule + batch\_mean \\
& Phase 2 checkpoint selection: lowest validation RMSE \\
& Phase 3 checkpoint selection: highest validation Macro-F1 \\
& Random seeds: 42 / 52 / 62 \\
\bottomrule[1pt]
\end{tabular}
\end{table*}
\section{Implementation Details}
\label{app:impl}
\paragraph{Module roles.}
The encoder, $\ffm$, and $\fds$ serve different purposes in \method{}.
The shared encoder backbone is instantiated twice, as $\EncImp$ for
cross-modal conditioning and as $\EncFus$ for downstream prediction.
In both cases, each modality is linearly projected, enriched with
temporal positional encoding, processed by two per-modality Informer
layers, and then fused across modalities by a cross-modal
MoE block with Informer attention~\cite{zhou2021informer}. The
resulting fused vector is used either as
the imputation context $\mathbf{c}_{\mathrm{imp}}$ or as the
downstream-task context $\mathbf{c}_{\mathrm{fus}}$.

$\ffm{} $is implemented as a conditional velocity network.
Given the interpolated state $x_\tau$, it combines a continuous FM time
embedding with side information built from time embeddings, feature
embeddings, the observation mask, and $\mathbf{c}_{\mathrm{imp}}$, then
predicts the velocity through four residual blocks with alternating
time- and feature-wise attention. During training, we sample one
continuous FM time $\tau \sim \mathcal{U}[0,1]$ independently for each
sample in the batch, construct $x_\tau = (1-\tau)x_0 + \tau x_1$, and
inject $\tau$ into the network through a sinusoidal continuous-time
embedding followed by a two-layer MLP. At inference time, we discretize
$[0,1]$ into 20 uniform Euler steps and evaluate the velocity field on
that fixed grid. Finally, $\fds$ is a lightweight
residual MLP head consisting of two linear projections with ReLU and
dropout, followed by a linear output layer that maps
$\mathbf{c}_{\mathrm{fus}}$ to the task label.

\paragraph{Hyperparameter.}
Table~\ref{tab:impl_hparams} summarizes the shared model and training
configurations used in all main experiments, including the common
three-phase schedule of 5/20/20 epochs.

\paragraph{Checkpoint selection.}
For the standard three-phase curriculum, Phase~2 saves
\texttt{best\_phase2\_diffusion.pt} and Phase~3 is initialized from that best
Phase~2 checkpoint rather than from the final epoch indiscriminately. Phase~3
then saves \texttt{best\_phase3\_joint.pt}, which is the checkpoint used for
all final test-set evaluation scripts. The selection criteria differ by stage:
the diffusion phase tracks the best validation RMSE because its goal is
signal-level recovery, whereas the final classification phase tracks the best
validation Macro-F1 on the downstream task. This two-stage selection rule keeps
the hand-off between imputation and classification aligned with the objective of
each phase instead of forcing one metric to govern the entire curriculum.

\paragraph{Training hardware.}
All experiments were trained on NVIDIA DGX B200 systems, using one NVIDIA
B200 GPU per run. For the common three-phase schedule of 5/20/20 epochs,
the representative wall-clock training time was approximately 55.2 minutes on
PPG-DaLiA (median over three runs), 9.8 hours on Sleep-EDF, 25.3 minutes on
PTB-XL, and 1.64 hours on Chapman-Shaoxing.

\section{Additional Results}
\label{app:additional}

This section collects appendix-only analyses that elaborate on design choices
supporting the main paper while remaining secondary to the core narrative. We
focus on secondary quantitative results that help interpret design choices and
complete the experimental picture, but that are not essential for understanding
the main method or its headline performance. We begin with a direct ablation of
the prior initialization strategies introduced in Section~\ref{app:prior}.

\subsection{Design Choice Ablation}
\label{app:prior_ablation}

Table~\ref{tab:prior_ablation} reports the effect of prior initialization
choices on downstream classification under the corresponding missingness type.
The goal of this ablation is not to maximize every appendix number with a more
complicated prior, but to test whether additional adaptivity consistently
translates into better downstream behavior. Because the proposed framework uses
the prior only as a source initialization for flow matching, a stronger prior
should ideally shorten the transport path without introducing instability or
overfitting. The table therefore helps distinguish improvements that are
structurally meaningful from those that merely add parameters.

\begin{table}[t]
\centering
\small
\caption{Ablation of prior initialization strategies on PTB-XL.
  \wmissing{} variants compare neighbor-rule initialization and
  learned-neighbor initialization under \wmissing{} 20\%.
  \mmissing{} variants compare batch-matched, batch-mean (default), and
  memory-bank initialization under \mmissing{} 20\%.
  Means and standard deviations over three runs are rounded to three decimals.
  Best per group in \textbf{bold}.}
\label{tab:prior_ablation}
\vspace{.5em}
\setlength{\tabcolsep}{30pt}
\begin{tabular}{llc}
\toprule[1pt]
\textbf{Missing type} & \textbf{Prior variant} & \textbf{Macro-F1} \\
\midrule

\multirow{2}{*}{\wmissing{}} & Neighbor rule
& $0.661_{\pm 0.003}$ \\
& Neighbor learned
& $\mathbf{0.668}_{\pm 0.007}$ \\
\midrule
\multirow{3}{*}{\mmissing{}} & Batch-matched
& $\mathbf{0.650}_{\pm 0.002}$ \\
& Batch-mean (default)
& $0.648_{\pm 0.002}$ \\
& Memory Bank
& $0.475_{\pm 0.002}$ \\
\bottomrule[1pt]
\end{tabular}
\end{table}

\noindent For the refreshed PTB-XL \mmissing{} reruns at $d_{\mathrm{modal}}=128$,
the corresponding AUROC values are $0.880_{\pm 0.001}$ for batch-matched and
$0.878_{\pm 0.002}$ for the default batch-mean prior. In contrast, the learned
memory-bank prior performs much worse, likely because a small set of learned
prototypes collapses toward overly averaged modality-level templates and loses
the sample-specific cross-modal structure needed for reliable initialization.

The \wmissing{} results suggest that adding limited adaptivity can be helpful
when the relevant uncertainty is still strongly local: the learned-neighbor
variant slightly outperforms the rule-based decay, which is consistent with the
idea that the model can refine how much trust to place in nearby observations
without abandoning the temporal-neighbor inductive bias. For \mmissing{}, the
picture is different. Batch-matched retrieval performs only marginally better
than the parameter-free batch mean, indicating that once an entire modality is
absent, coarse cross-sample context is useful but does not dramatically change
the downstream outcome. The large drop of the memory-bank variant further
suggests that introducing learnable global prototypes is not automatically
beneficial: if those prototypes fail to preserve sample-specific cross-modal
alignment, they can become worse initial anchors than even a simple global
average.

\noindent\textbf{Default configuration.}
In the main experiments, we use the rule-based within-modality prior and the
batch-mean modality-level prior.
Appendix~\ref{app:prior_ablation} shows that batch-matched retrieval is only
slightly stronger than the default batch mean, whereas the learned memory-bank
variant performs substantially worse. We therefore keep the parameter-free
batch-mean prior in the primary setting because it remains competitive without
introducing additional learned prototypes. More broadly, these results support
our main design principle that prior-aware initialization should encode the
right missingness structure with as little unnecessary complexity as possible.

\subsection{Exact HPO Values Used in Figure~\ref{fig:hpo_curves}}
\label{app:hpo_values}

Table~\ref{tab:hpo_values} lists the paper-facing mean and standard deviation
values used to generate Figure~\ref{fig:hpo_curves}. We report these values
directly from the plotting source so that the appendix matches the figure
exactly, even though the broader HPO archive in \texttt{results\_overall.typ} still
contains historical entries that are not fully synchronized with the final
paper-facing plot.

\begin{table}[t]
\centering
\small
\caption{Exact HPO values used in Figure~\ref{fig:hpo_curves} on PTB-XL under
  mixed 20\% missingness. The exploratory $N{=}1$ point is omitted to match the
  plotted figure. Means and standard deviations are reported exactly from the
  paper-facing plotting source.}
\label{tab:hpo_values}
\vspace{.5em}
\setlength{\tabcolsep}{32pt}
\begin{tabular}{llcc}
\toprule[1pt]
\textbf{Sweep} & \textbf{Setting} & \textbf{Macro-F1} & \textbf{AUROC} \\
\midrule
\multicolumn{4}{l}{\textit{Number of experts $N$}} \\
$N$ & $2$  & $0.618_{\pm 0.006}$ & $0.850_{\pm 0.005}$ \\
$N$ & $4$  & $\mathbf{0.641}_{\pm 0.011}$ & $\mathbf{0.867}_{\pm 0.000}$ \\
$N$ & $8$  & $0.622_{\pm 0.010}$ & $0.861_{\pm 0.002}$ \\
$N$ & $12$ & $0.616_{\pm 0.007}$ & $0.848_{\pm 0.002}$ \\
$N$ & $16$ & $0.605_{\pm 0.026}$ & $0.851_{\pm 0.005}$ \\
$N$ & $32$ & $0.617_{\pm 0.008}$ & $0.847_{\pm 0.004}$ \\
\midrule
\multicolumn{4}{l}{\textit{Expert size $d_{\mathrm{modal}}$}} \\
$d_{\mathrm{modal}}$ & $16$  & $0.586_{\pm 0.011}$ & $0.846_{\pm 0.004}$ \\
$d_{\mathrm{modal}}$ & $32$  & $0.622_{\pm 0.008}$ & $0.851_{\pm 0.002}$ \\
$d_{\mathrm{modal}}$ & $64$  & $0.612_{\pm 0.022}$ & $0.842_{\pm 0.005}$ \\
$d_{\mathrm{modal}}$ & $128$ & $\mathbf{0.641}_{\pm 0.011}$ & $\mathbf{0.867}_{\pm 0.000}$ \\
$d_{\mathrm{modal}}$ & $256$ & $0.311_{\pm 0.132}$ & $0.786_{\pm 0.017}$ \\
\bottomrule[1pt]
\end{tabular}
\end{table}

\noindent The table makes the two trends in Figure~\ref{fig:hpo_curves}
explicit. Increasing the number of experts beyond $N{=}4$ does not deliver a
stable gain, and overly large expert dimensionality quickly becomes harmful:
performance improves from $d_{\mathrm{modal}}{=}16$ to $128$, but collapses at
$256$. These exact values support the final choice of $N{=}4$ and
$d_{\mathrm{modal}}{=}128$ in the main configuration.

\subsection{Clean-Input Performance}
\label{app:clean_results}

Table~\ref{tab:clean_results} reports clean-input performance on the three
benchmarks for which we currently maintain complete clean summaries in the
benchmark archive: Sleep-EDF, PTB-XL, and PPG-DaLiA. We include this table to
show that the missing-aware design of \method{} does not depend on evaluating
only corrupted inputs. To keep this appendix table aligned with the main-paper
comparison set, we report the same primary baselines as in the main results and
do not include ShaSpec or CLIMB here. Even without synthetic missingness, the
model remains competitive and achieves the best Macro-F1 on Sleep-EDF and
PTB-XL within this primary comparison set, while remaining close to the
strongest baseline on PPG-DaLiA.

\begin{table*}[t]
\centering
\small
\caption{Clean-input classification performance on the three benchmarks with
  complete clean summaries in the current archive. To match the main-paper
  comparison set, the table reports the same primary baselines and excludes
  ShaSpec and CLIMB. Means and standard deviations over three runs are rounded
  to three decimals. \textbf{Bold}: best; \underline{underlined}: second-best
  within the reported comparison set.}
\label{tab:clean_results}
\vspace{.5em}
\setlength{\tabcolsep}{6pt}
\begin{tabular}{l cc cc cc}
\toprule[1pt]
& \multicolumn{2}{c}{\textbf{Sleep-EDF}}
& \multicolumn{2}{c}{\textbf{PTB-XL}}
& \multicolumn{2}{c}{\textbf{PPG-DaLiA}} \\
\cmidrule(lr){2-3}\cmidrule(lr){4-5}\cmidrule(lr){6-7}
\textbf{Method} & F1 & AUROC & F1 & AUROC & F1 & AUROC \\
\midrule
FuseMoE & $0.582_{\pm 0.004}$ & $0.930_{\pm 0.001}$ & $0.224_{\pm 0.034}$ & $0.686_{\pm 0.023}$ & $0.623_{\pm 0.019}$ & $0.925_{\pm 0.002}$ \\
Flex-MoE & $\underline{0.704}_{\pm 0.011}$ & $\underline{0.973}_{\pm 0.002}$ & $0.440_{\pm 0.016}$ & $0.767_{\pm 0.004}$ & $0.624_{\pm 0.004}$ & $0.928_{\pm 0.001}$ \\
Maestro & $0.699_{\pm 0.004}$ & $0.971_{\pm 0.002}$ & $\underline{0.622}_{\pm 0.009}$ & $\underline{0.882}_{\pm 0.002}$ & $\mathbf{0.769}_{\pm 0.004}$ & $\mathbf{0.958}_{\pm 0.002}$ \\
MIRA & $0.629_{\pm 0.000}$ & $0.958_{\pm 0.001}$ & $0.482_{\pm 0.022}$ & $0.826_{\pm 0.006}$ & $0.676_{\pm 0.004}$ & $0.939_{\pm 0.002}$ \\
\method{} & $\mathbf{0.744}_{\pm 0.007}$ & $\mathbf{0.981}_{\pm 0.002}$ & $\mathbf{0.657}_{\pm 0.005}$ & $\mathbf{0.884}_{\pm 0.002}$ & $\underline{0.748}_{\pm 0.011}$ & $\underline{0.951}_{\pm 0.003}$ \\
\bottomrule[1pt]
\end{tabular}
\end{table*}

\noindent These clean results provide a useful complement to the missing-input
tables in the main paper. Within the primary baseline set used throughout the
main text, \method{} remains the strongest method on Sleep-EDF and PTB-XL even
without injected missingness, suggesting that the shared encoder and
imputed-data fine-tuning strategy do not trade away clean-case classification
quality. On PPG-DaLiA, Maestro remains the strongest baseline, but \method{}
still ranks second on both Macro-F1 and AUROC within the same comparison set.

\newpage

\end{document}